\documentclass[acmsmall]{acmart}

\AtBeginDocument{%
  \providecommand\BibTeX{{%
    \normalfont B\kern-0.5em{\scshape i\kern-0.25em b}\kern-0.8em\TeX}}}

\setcopyright{acmlicensed}
\copyrightyear{2018}
\acmYear{2018}
\acmDOI{XXXXXXX.XXXXXXX}

\acmJournal{TIST}
\acmVolume{37}
\acmNumber{4}
\acmArticle{28}
\acmMonth{8}
\usepackage{multirow}

\begin{document}

\title{Survey on Knowledge Distillation for Large Language Models: Methods, Evaluation, and Application}

\author{Chuanpeng Yang}
\affiliation{%
  \institution{
Institute of Information Engineering,
Chinese Academy of Sciences \&  
School of Cyber Security, University
of Chinese Academy of Sciences}
\city{Beijing}
\country{China}
}
\orcid{0000-0003-0601-2224}
\email{yangchuanpeng@iie.ac.cn}

\author{Wang Lu}
\authornote{Correspondence to: Wang Lu and Yao Zhu.}
\affiliation{%
  \institution{Tsinghua University}
  \country{China}}
\email{newlw230630@gmail.com}

\author{Yao Zhu}
\authornotemark[1]
\affiliation{%
  \institution{Zhejiang University}
  \country{China}}
\email{ee_zhuy@zju.edu.cn}

\author{Yidong Wang}
\affiliation{%
  \institution{Peking University}
  \city{Haidian Qu}
  \state{Beijing Shi}
  \country{China}
}
\email{yidongwang37@gmail.com}

\author{Qian Chen}
\affiliation{%
  \institution{Institute of Computing Technology, Chinese Academy of Sciences, University of Chinese Academy of Sciences}
  \country{China}
}
\email{chenqian20b@ict.ac.cn}

\author{Chenlong Gao}
\affiliation{%
  \institution{Institute of Computing Technology, Chinese Academy of Sciences}
  \country{China}
}
\email{gaochenlong@ict.ac.cn}

\author{Bingjie Yan}
\affiliation{%
  \institution{Institute of Computing Technology, Chinese Academy of Sciences, University of Chinese Academy of Sciences}
  \country{China}
}
\email{bj.yan@ieee.org}

\author{Yiqiang Chen}
\affiliation{%
  \institution{Institute of Computing Technology, Chinese Academy of Sciences}
  \country{China}
}
\email{yqchen@ict.ac.cn}

\renewcommand{\shortauthors}{Chuanpeng Yang, et al.}
\newcommand{\lw}[1]{{\color{blue}{[LW: #1]}}}
\begin{abstract}
Large Language Models (LLMs) have showcased exceptional capabilities in various domains, attracting significant interest from both academia and industry. Despite their impressive performance, the substantial size and computational demands of LLMs pose considerable challenges for practical deployment, particularly in environments with limited resources. The endeavor to compress language models while maintaining their accuracy has become a focal point of research. Among the various methods, knowledge distillation has emerged as an effective technique to enhance inference speed without greatly compromising performance. This paper presents a thorough survey from three aspects: method, evaluation, and application, exploring knowledge distillation techniques tailored specifically for LLMs. Specifically, we divide the methods into white-box KD and black-box KD to better illustrate their differences. Furthermore, we also explored the evaluation tasks and distillation effects between different distillation methods, and proposed directions for future research. Through in-depth understanding of the latest advancements and practical applications, this survey provides valuable resources for researchers, paving the way for sustained progress in this field.
\end{abstract}



\received{20 February 2007}
\received[revised]{12 March 2009}
\received[accepted]{5 June 2009}

\newcommand{\etal}{\textit{et al}.}
\newcommand{\ie}{\textit{i}.\textit{e}.}
\newcommand{\eg}{\textit{e}.\textit{g}.}
\maketitle

\section{Introduction}

\begin{figure*}
	\centering{\includegraphics[width=\textwidth]{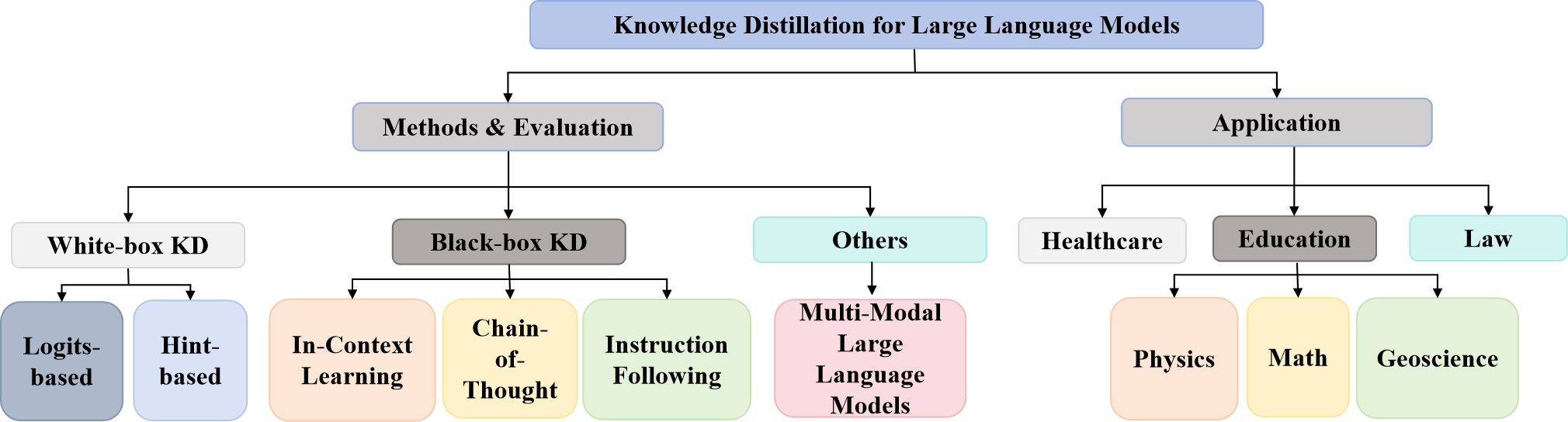}}
	\caption{The general taxonomy framework of this survey.}
	\label{fig1}
\end{figure*}

The emergence of Large Language Models (LLMs) \cite{touvron2023llama, achiam2023gpt, zhao2023survey, chang2023survey, wang2023pandalm} has significantly improved text generation quality in various generative tasks, becoming a pivotal and widely discussed topic in the field of artificial intelligence. These models, compared to their predecessors, show superior generalization to unseen data. Moreover, they exhibit capabilities that smaller models lack, such as multi-step reasoning \cite{li2022explanations, magister2022teaching, hsieh2023distilling} and instruction-following \cite{wang2022self, peng2023instruction, wu2023lamini}. The success of LLMs is often attributed to increased training data and a larger number of model parameters (e.g., GPT-3 with 175 billion parameters \cite{brown2020language}). However, the expansion in parameter size brings significant drawbacks, particularly in terms of high inference costs and substantial memory requirements, making practical deployment challenging. For example, GPT-3 requires around 350GB of model storage (float16) and at least 5 A100 GPUs with 80GB of memory each for inference, contributing significantly to carbon emissions. To mitigate these challenges, model compression \cite{deng2020model, he2018amc} has emerged as a viable solution. Model compression aims to transform large, resource-heavy models into more compact versions that are suitable for storage on constrained mobile devices. This process may involve optimizing for reduced latency to achieve faster execution or balancing between minimal latency and model performance. Thus, a key goal in applying these high-capacity models in real-world scenarios is to compress them, reducing the number of parameters while preserving maximum performance.

As the necessity to reduce computational resource demands becomes increasingly crucial, Knowledge Distillation (KD) \cite{hinton2015distilling} emerges as a promising technique. KD is a machine learning method focused on compressing and speeding up models by transferring knowledge from a large, complex model to a smaller, more efficient one. This technique is frequently employed to condense the knowledge stored in large deep neural network models into smaller counterparts, thus reducing computational resource requirements and improving inference speed without substantial performance sacrifices. Fundamentally, knowledge distillation leverages the extensive knowledge acquired by a large model on a substantial dataset to guide the training of a smaller model. This knowledge typically includes the output probability distribution, intermediate layer representations, and loss function of the large model. During training, the smaller models aim not only to match the original data labels but also to mimic the behavior of the larger models. For advanced models like GPT-4 \cite{achiam2023gpt}, which are accessible only through APIs, the generated instructions and explanations can aid in the training of student models \cite{jiang2023lion}.

With recent advancements in knowledge distillation, several studies have synthesized the latest progress in various distillation techniques. Specifically,
Gou \etal \cite{gou2021knowledge} provide an extensive review of knowledge distillation, addressing six critical aspects: knowledge categories, training schemes, teacher-student architectures, distillation algorithms, performance comparisons, and applications. Similarly, Wang \etal \cite{wang2021knowledge} summarize the research progress and technical details of knowledge distillation techniques related to visual tasks comprehensively. Alkhulaifi \etal \cite{alkhulaifi2021knowledge} introduce an innovative metric known as the distillation metric, which they employ to evaluate different knowledge compression methods. Additionally, Hu \etal \cite{hu2023teacher} explore various teacher-student architectures across multiple distillation objectives, presenting different knowledge representations and their corresponding optimization goals. They also provide a systematic overview of teacher-student architectures, incorporating representative learning algorithms and effective distillation schemes.

Existing reviews on knowledge distillation have laid a crucial foundation and offered valuable insights into model compression \cite{calderon-etal-2023-systematic,lee-etal-2023-study,10.1109/TASLP.2023.3289303}. However, the emergence of LLMs has brought several new challenges to KD: 1) Large language models are designed not for single tasks like text generation but for broad generality across various tasks and unseen data, including emergent capabilities. Therefore, assessing the generalization of compressed LLMs requires careful and thorough evaluation. 2) The existing review is only a summary of existing work, without providing specific examples of KD technology applied to compress and deploy LLMs in real-world scenarios. This case study can help readers choose the best KD scheme for LLMs of different scales.

To tackle these challenges, a variety of knowledge distillation algorithms specifically designed for LLMs have been developed. This paper aims to provide a comprehensive and insightful guide to these methods. The overarching classification framework of our survey is depicted in Figure \ref{fig1}, which examines the distillation algorithm of LLMs from three aspects: method, evaluation, and application. To clearly explain these methods, we categorize them into white-box KD and black-box KD. White-box KD includes two distinct types: Logits-based methods \cite{hinton2015distilling}, which transfer knowledge at the logits level, and Hint-based methods \cite{romero2014fitnets}, which transmit knowledge through intermediate features. Black-box KD involves an API-based approach where only the outputs from the teacher model are accessible. This category typically includes three methods: In-Context Learning \cite{huang2022context}, Chain-of-Thought \cite{li2022explanations}, and Instruction Following \cite{wang2022self}. In addition, we simultaneously evaluate the effectiveness of the above two types of distillation algorithms on robustness benchmarks \cite{nie-etal-2020-adversarial,wang2021adversarial,tchango2022ddxplus}. Finally, we discuss the relationships and application scenarios among different distillation methods and propose directions for future research.

The rest of this paper is organized as follows. Sec.\ref{2} briefly reviews the definition of knowledge distillation methods. Next, Sec.\ref{3} delves into the distillation and evaluation methods in the field of LLMs. Sec.\ref{sec-app} presents applications while Sec.\ref{4} summarizes the challenges of knowledge distillation and explores future research directions. Lastly, Sec.\ref{5} concludes the paper.


\section{Overview of Knowledge Distillation }
\label{2}
In this section, we summarized the optimization objectives of each knowledge distillation algorithm.

\subsection{Logits-based KD}

As implied by its name, logic-based KD \cite{hinton2015distilling} is a distillation paradigm that employs logic within teacher models for knowledge transfer. We can formulate the general knowledge distillation loss function as follows:
\begin{equation}
\mathcal{L}_{\text {logits }}=K L\left(p^{\mathbf{t}} \| p^{\mathbf{s}}\right)=\sum_{j=1}^C p_j^{\mathbf{t}} \log \left(\frac{p_j^{\mathbf{t}}}{p_j^{\mathbf{s}}}\right),
\end{equation}
\begin{equation}
p_i^{\mathbf{s}}=\frac{\exp \left(z_i^{\mathbf{s}}/\tau\right)}{\sum_{j=1}^C \exp \left(z_j^{\mathbf{s}}/\tau\right)}, \quad p_i^{\mathbf{t}}=\frac{\exp \left(z_i^{\mathbf{t}}/\tau\right)}{\sum_{j=1}^C \exp \left(z_i^{\mathbf{t}}/\tau\right)},
\end{equation}
where $z^{\mathbf{s}}$, $z^{\mathbf{t}} \in \mathbb{R}^C$ denote the logits output of the student and teacher network, respectively. $\tau$ is a temperature parameter that adjusts the smoothness of the logits. $C$ represents the number of classes. The Kullback-Leibler divergence (KLD) \cite{hinton2015distilling} loss can also be replaced with other functions, such as Reverse Kullback–Leibler (RKL) \cite{huszar2015not, nowozin2016f, chen2018symmetric, lee2023self} distillation, Jenson–Shannon (JS) \cite{tian2021farewell} distillation, etc.

\subsection{Hint-based KD}

Given the restricted ability of students to extract knowledge in logit-based knowledge distillation, researchers strive to more precisely replicate the behavior of teachers.  
Consequently, intermediate feature-based knowledge distillation \cite{sun2019patient, hou2020dynabert} was introduced. This technique involves matching the outputs of the intermediate layers between student and teacher models. This approach requires students to understand both the results and the processes leading to those results. The general form of the feature-based knowledge distillation loss function is outlined below:
\begin{equation}
\mathcal{L}_{\text {hint }}=\mathcal{H}\left(F^{\mathbf{s}}, F^{\mathbf{t}}\right)=\left\|F^{\mathbf{t}}-\phi\left(F^{\mathbf{s}}\right)\right\|^2,
\end{equation}
where $F^{\mathbf{s}}, F^{\mathbf{t}}\in \mathbb{R}^{H\times W \times C}$ denote the intermediate features of the student and teacher networks, respectively. The function $\phi$ is used to ensure that the student features match the dimensions of the teacher features. The metric function is represented by $\mathcal{H}$, and as an example, we use mean square error.

\subsection{In-Context Learning}
ICL \cite{brown2020language,huang2022context} utilizes a natural language prompt composed of task descriptions or several task examples as demonstrations. Formally, let $D_K=\{ f\left(x_1, y_1\right), \ldots, f\left(x_k, y_k\right.) \} $ represent a set of $k$ examples, where $f\left(x_k, y_k\right.)$ is a function that converts the $k$-th task example into a natural language prompt. Given the task description $I$, the demonstration set $D_k$, and a new input query $x_{k+1}$, the predicted output $\hat{y}_{k+1}$ generated by the LLM can be described by the following formula:
\begin{equation}
\operatorname{LLM}(I, \underbrace{f\left(x_1, y_1\right), \ldots, f\left(x_k, y_k\right.}_{\text {demonstrations }}), f(\underbrace{x_{k+1}}_{\text {input }}, \underbrace{\underline{ }}_{\text {answer }})) \rightarrow \hat{y}_{k+1} \text {,}
\end{equation}
where the answer $y_{k+1}$ is left blank for the LLM to predict. The student model is used to predict the results generated by the LLM.

\subsection{Chain-of-Thought}

CoT \cite{li2022explanations, magister2022teaching, hsieh2023distilling, wadhwa2023revisiting} integrates intermediate reasoning steps into prompts, rather than relying solely on simple input-output pairs as done in ICL.
\begin{equation}
\operatorname{LLM}(I, \underbrace{f\left(x_1, r_1, y_1\right), \ldots, f\left(x_k, r_k, y_k\right.}_{\text {demonstrations }}), f(\underbrace{x_{k+1}}_{\text {input }}, \underbrace{\underline{ }}_{\text {rational }},\underbrace{\underline{ }}_{\text {answer }})) \rightarrow \hat{r}_{k+1}, \hat{y}_{k+1} \text {. }
\end{equation}
where $r_k$ represents the rationale provided by the user that explains why the answer to $x_k$ is $y_k$. At this point, the student model not only needs to predict the labels of the teacher model, but also needs to emulate the reasons generated by the teacher. 

\subsection{Instruction Following}

By fine-tuning on a structured multitask dataset that utilizes natural language descriptions, LLMs exhibit proficiency on unseen tasks that are similarly expressed in instructional formats \cite{sanh2021multitask, ouyang2022training, wei2021finetuned}. Through instruction tuning, LLMs can follow task guidelines for new assignments without needing explicit examples, thus improving their generalization abilities. The process of distilling instruction-following skills involves generating task-specific instructions with the LLM and then fine-tuning the student model using this instruction dataset.

\section{Knowledge Distillation in Large Language Models}
\label{3}
The Transformer architecture is highly scalable, allowing for the creation of extremely large models with billions or even trillions of parameters. It underpins many of the most prominent large-scale models in NLP, CV, and multimodal domains.  For example, notable large language models like the GPT series \cite{brown2020language, achiam2023gpt}, LLaMA \cite{touvron2023llama}, and Qwen \cite{bai2023qwen} are based on its decoder-only configuration. Before 2023, research on Transformer-based NLP distillation \cite{tang2019distilling, sanh2019distilbert} mainly centered around the BERT architecture. However, with the rise of pre-trained large language models \cite{ouyang2022training, achiam2023gpt}, there has been increasing interest in distilling Transformers with billion-scale parameters and in developing more efficient distillation methods for scenarios with limited data and high computational costs \cite{ho2022large, hsieh2023distilling}. The existing distillation algorithms are mainly divided into two categories: white-box KD and black-box KD.

\subsection{White-box Knowledge Distillation}

White-box distillation depends on methods that require access to the teacher model's internal data during training, utilizing the accessible internal information of the teacher model. In the following discussion, we explore two distinct types of white-box knowledge distillation. Firstly, logits-based methods, introduced by Hinton \etal \cite{hinton2015distilling}, transfer knowledge at the logits level, where the knowledge is conveyed using the teacher model's logits. Given the limited knowledge acquired by students in logits-based knowledge distillation, researchers aim to more accurately replicate the teacher's behavior. To this end, Romero \etal \cite{romero2014fitnets} propose hint-based knowledge distillation, which involves aligning the feature outputs of intermediate layers between the student and teacher models. This approach requires the student to understand not only the final results but also the processes leading to those results. In the following section, we analyze in detail the characteristics of each method from the perspective of evaluation tasks (as shown in Table \ref{tab2}). Furthermore, we evaluate the strengths and weaknesses of the two types of distillation algorithms based on robustness, providing certain guidance in the applicable scenarios of the algorithms.

\begin{table}
\caption{Comparison of representative white-box KD methods on large language models. The compression ratio indicates the proportion of the original uncompressed model size to the compressed model size.}
\label{tab2}
\resizebox{\linewidth}{!}{
\begin{tabular}{@{}llcccc@{}}
\toprule
Models & Distillation Type  & Teacher Model  & Compression Rate  & Evaluation Task & Comparison with Teacher Model

\\ \midrule
DistillBiLSTM \cite{tang2019distilling}    & logits-based  &  BERT$_{base}$  & 114$\times$   & GLUE\cite{wang2018glue}: SST-2/QQP/MNLI-m/MNLI-mm  &  81.1/87.7 (92\% performance) \\  
DistillBERT \cite{sanh2019distilbert}    & logits-based  &  BERT$_{base}$  & 2$\times$   &  GLUE\cite{wang2018glue}: SST-2/QQP/MNLI/QNLI/RTE/MRPC/CoLA/STS-B/WNLI  &  77.0/79.5 (97\% performance)   \\ 
MixKD \cite{liang2020mixkd}    & logits-based  &  BERT$_{base}$  & 2$\times$   &  GLUE\cite{wang2018glue}: SST-2/QQP/MNLI-m/MNLI-mm/QNLI/RTE/MRPC  &  77.2/82.6 (93\% performance) \\
ReAugKD \cite{zhang-etal-2023-reaugkd}    & logits-based  &  BERT$_{base}$ & 2$\times$   &  GLUE\cite{wang2018glue}: SST-2/QQP/QNLI/RTE/MRPC/CoLA  &  81.8/82.3 (99\% performance)  \\ 
PD \cite{turc2019well}    & logits-based  &  BERT$_{base}$  & 2$\times$     &  GLUE\cite{wang2018glue}: SST-2/QQP/MNLI/QNLI/RTE/MRPC  &  82.1/81.7 (100.5\% performance) \\ \midrule
MINILLM \cite{gu2023knowledge}   & logits-based  & \begin{tabular}[c]{@{}l@{}}GPT-2$_{1.5B}$ \\ OPT$_{13B}$ \\ LLaMA$_{13B}$\end{tabular} & \begin{tabular}[c]{@{}l@{}} 12$\times$ \\ 10$\times$ \\ 2$\times$ \end{tabular} &  Dolly\cite{gu2023knowledge}/SelfInst\cite{wang2022self}/Vicuna\cite{chiang2023vicuna}/S-NI\cite{wang2022benchmarking}/UnNI\cite{honovich2022unnatural}  &   \begin{tabular}[c]{@{}l@{}} 22.0 \textasciitilde 25.5 / 24.1  (91\% \textasciitilde 106\% performance) \\  24.3 \textasciitilde 26.9 / 29.4  (92\% \textasciitilde 102\% performance) \\ 29.7 / 29.4  (101\% performance) \end{tabular} \\ \midrule
GKD \cite{agarwal2023generalized}    & logits-based  &  T5$_{XL}$  & 39$\times$  &  XSum\cite{narayan2018don}/WMT14EN-DE\cite{bojar2014findings}/GSM8K\cite{cobbe2021training}/MMLU\cite{hendrycks2020measuring}/BBH\cite{srivastava2023beyond}  & 14.5/26.0 (56\% performance) \\ \midrule
MiniMA \cite{zhang2023towards} & logits-based  &  LLaMA-2$_{7B}$  & 2$\times$   &  MMLU\cite{hendrycks2020measuring}/CEval\cite{tunstall2023zephyr}/DROP\cite{dua2019drop}/BBH\cite{srivastava2023beyond}/GSM8K\cite{cobbe2021training} HumanEval\cite{chen2021evaluating} & 21.7/28.5 (76\% performance) \\ \midrule
PKD \cite{sun2019patient}   & hint-based  &  BERT$_{base}$  & 2$\times$   & GLUE\cite{wang2018glue}: SST-2/QQP/MNLI-m/MNLI-mm/QNLI/RTE/MRPC & 77.7/84.9 (92\% performance)   \\  
MetaDistil \cite{zhou2022bert}    & hint-based  &  BERT$_{base}$  & 2$\times$    &  GLUE\cite{wang2018glue}: SST-2/QQP/MNLI/QNLI/RTE/MRPC/CoLA/STS-B  & 80.4/80.7 (99\% performance) \\ 
AD-KD \cite{wu-etal-2023-ad}    & hint-based  &  BERT$_{base}$  & 2$\times$   &  GLUE\cite{wang2018glue}: SST-2/QQP/MNLI/QNLI/RTE/MRPC/CoLA/STS-B & 83.4/84.1 (99\% performance)   \\ 
XtremeDistil \cite{mukherjee}    & hint-based  &  mBERT$_{base}$  & 35$\times$  &  Multilingual NER\cite{pan}/IMDB\cite{learning}/SST-2\cite{socher}/Elec\cite{2507163}/DbPedia\cite{NIPS2015_250cf8b5}/Ag News\cite{NIPS2015_250cf8b5} & 88.6/92.7 (95\% performance)    \\ 
TinyBERT \cite{tinybert}    & hint-based  &  BERT$_{base}$  & 7$\times$   &  GLUE\cite{wang2018glue}: SST-2/QQP/MNLI-m/MNLI-mm/QNLI/RTE/MRPC/CoLA/STS-B  & 77.0/79.5 (97\% performance)  \\
MobileBERT \cite{mobilebert}    & hint-based  &  IB-BERT$_{large}$  & 4$\times$   &  GLUE\cite{wang2018glue}: SST-2/QQP/MNLI-m/MNLI-mm/QNLI/RTE/MRPC/CoLA/STS-B & 77.7/78.3 (99\% performance)  \\
MiniLM \cite{496209}    & hint-based  &  BERT$_{base}$  & 2$\times$  &  SQuAD2\cite{Rajpurkar_Jia_Liang_2018}/ GLUE\cite{wang2018glue}: SST-2/MNLI-m & 80.4/81.5 (99\% performance) \\
TED \cite{3619267}  & hint-based  &  DeBERTaV3$_{base}$  & 2$\times$   &  GLUE\cite{wang2018glue}:SST-2/QQP/MNLI-m/MNLI-mm/QNLI/RTE/MRPC/CoLA/STS-B  & 87.5/88.9 (98\% performance) \\
HomoDistil \cite{liang2023homodistil}  & hint-based  &  BERT$_{base}$  & 7$\times$  &  GLUE\cite{wang2018glue}: SST-2/QQP/MNLI/QNLI/RTE/MRPC/CoLA/STS-B  & 79.0/84.6 (93\% performance)  \\
\bottomrule
\end{tabular}
}
\end{table}

\subsubsection{Logits-based KD}  

The distillation of Bidirectional Long Short-Term Memory Networks (BiLSTM) \cite{tang2019distilling} marks the earliest attempt to apply knowledge distillation to BERT \cite{kenton2019bert}. The distillation objective is to minimize the mean squared error loss between the logits of the student network and those of the teacher. This approach has been tested on three tasks: sentence classification and sentence matching. Experimental results show that the shallow BiLSTM-based model achieves performance comparable to the ELMo language model \cite{peters2018deep}, but with approximately 100 times fewer parameters and a 15-fold increase in inference speed. Similarly, DistillBERT \cite{sanh2019distilbert} initializes a shallower student model using the teacher's parameters and minimizes the difference in soft target probabilities between the teacher and student, a technique known as word-level knowledge distillation. It introduced a triple loss that combines language modeling, distillation, and cosine distance loss to leverage the inductive bias learned by the pre-trained model. DistilBERT achieved performance equivalent to or exceeding the ELMo baseline in nine tasks. Compared to BERT, DistilBERT maintains 97\% of the performance while reducing the number of parameters by 40\%. MixKD \cite{liang2020mixkd} extends the concept of encouraging students to mimic teachers' logits by using linear interpolation of example pairs. It improves the effectiveness of knowledge distillation by using data augmentation to create additional samples from the available task-specific data. This approach mirrors students learning more effectively from teachers by asking further questions to explore their answers and concepts in depth, providing more data for student models to extract insights from large-scale language models. Evaluation results across six datasets show that MixKD significantly outperforms traditional knowledge distillation and previous methods in compressing large language models. ReAugKD \cite{zhang-etal-2023-reaugkd} includes both an inference stage and a training stage. In the inference stage, it aggregates soft labels generated by teachers that closely resemble student embeddings. During the training phase, a novel relationship KD loss is used to minimize the differences between teacher-student embeddings and their distributions. Evaluation results on six datasets demonstrated that ReAugKD achieved superior performance compared to the baseline, with a latency overhead of less than 3\% of the baseline, highlighting that integrating retrieval information can significantly improve generalization ability. Turc \etal \cite{turc2019well} proposed a pre-training distillation (PD) method, which is a universal yet straightforward algorithm for building compact models. It consists of three standard training operation sequences and can be applied to any architecture choice. The method also explores transferring task knowledge from large fine-tuned models using traditional logits-based KD and evaluates its performance on six datasets. On average, this pre-training distillation method performs best and even surpasses the corresponding teacher model. The above distillation algorithms are all based on BERT as the teacher model and GLUE as the evaluation benchmark. With the increasing size of the model, existing distillation algorithms and evaluation standards can no longer meet the requirements

MINILLM \cite{gu2023knowledge} addresses the limitations of traditional logits-based Knowledge Distillation methods by proposing an innovative approach to distill large language models (LLMs) into smaller ones, focusing on minimizing the forward Kullback-Leibler divergence during free-running generation. This method replaces the standard KD method's forward KLD target with a reverse KLD, which is more suitable for generating KD on language models and aims to prevent student models from overestimating the low probability distribution of teacher distributions. To further stabilize and accelerate training, an effective optimization method is introduced, comprising three key steps: 1) single-step decomposition to reduce variance, 2) teacher mixed sampling to mitigate reward hacking, and 3) length normalization to counteract length bias. MINILLM is applied to models ranging in size from 120M to 13B parameters. Experimental evaluations on five datasets using Rouge-L \cite{lin2004rouge}, human judgment, and GPT-4 feedback consistently demonstrate that this approach outperforms the standard KD baseline. Further research and analysis indicate that MINILLM can reduce exposure bias and improve long-response generation performance. Similar to MINILLM, GKD \cite{agarwal2023generalized} moves beyond relying solely on a fixed set of output sequences, training student models to generate their own sequences with feedback from the teacher model. Unlike supervised KD methods, GKD allows for the use of alternative loss functions between the student and teacher, which is advantageous when student models lack the expressive capability to effectively mimic teacher distributions. Additionally, GKD enables the seamless integration of distillation and Reinforcement Learning (RL) fine-tuning for language models. By providing flexibility to optimize alternative divergence measures such as reverse KL and generalized JSD, GKD allows limited student capacity to focus on generating samples similar to those produced under teacher supervision. It has been demonstrated that on-policy GKD facilitates the integration of distillation with RL \cite{ouyang2022training} fine-tuning of language models, a combination not previously explored. Regarding performance enhancement for initial students, on average, GKD yielded a relative gain of 2.1 times for abstracts, 1.7 times for machine translation, and 1.9 times for arithmetic reasoning tasks across different sizes of T5 student models, underscoring the effectiveness of GKD. In terms of performance enhancement for initial students, GKD showed average relative gains of 2.1 times for abstracts, 1.7 times for machine translation, and 1.9 times for arithmetic reasoning tasks across various sizes of T5 student models, highlighting the effectiveness of GKD. Wen \etal \cite{wen2023f} proposed the $f$-DISTILL framework, which formulates sequence-level knowledge distillation by minimizing a generalized $f$-divergence function. This framework introduces four distillation variants, demonstrating that existing SeqKD \cite{kim2016sequence} and ENGINE \cite{tu2020engine} methods are approximations of KL and reverse KL distillation. Furthermore, the $f$-DISTILL method includes step-wise decomposition to convert the complex sequence-level divergence into a more manageable word-level loss. This facilitates easier calculation. This method was evaluated on four datasets: DART for data-to-text generation \cite{nan2021dart}, XSum for summarization \cite{narayan2018don}, WMT16 EN-RO for machine translation \cite{bojar2016findings}, and Commonsense Dialogue \cite{zhou2021commonsense}. The experiments demonstrated that $f$-DISTILL variants outperformed existing distribution-matching KD methods, leading to performance improvements when combined with representation-matching KD methods.Additionally, the results indicated that symmetric distillation loss is superior to asymmetric distillation loss, confirming that extreme mode averaging or collapse is suboptimal. MiniMA \cite{zhang2023towards} found that the optimal distillation effect occurs when the student model is approximately 40\% the size of the teacher model. It combines structured pruning with logit-based knowledge distillation, using LLaMA2-7B \cite{touvron2023llama} as the teacher model to train the 3B MiniMA model. The results showed that MiniMA achieved impressive performance in knowledge, reasoning, and encoding, while using a similar or even fewer number of tokens than the teacher model.


\subsubsection{Hint-based KD}  

The feature-based knowledge distillation methods \cite{sun2019patient, hou2020dynabert} extract knowledge from the embedding space, transformer layers, and prediction layers, allowing the student model to learn various aspects of the teacher model comprehensively. For instance, Sun \etal \cite{sun2019patient} proposed a patient knowledge distillation (PKD) method aimed at compressing a large-scale teacher model into an equally effective lightweight student model. They proposed two distinct distillation strategies: 1) PKD-Last: The student model learns from the last $k$ layers of the teacher model, based on the assumption that the top layers contain the most informative knowledge. 2) PKD-Skip: The student learns from every $k$-layer of the teacher, suggesting that the lower layers also contain essential information that should be gradually transferred during distillation. Experiments conducted on seven datasets across four tasks—sentiment classification, paraphrase similarity matching, natural language inference, and machine reading comprehension—showed that the PKD method outperformed standard knowledge distillation methods. It achieved superior performance and better generalization, significantly enhancing training efficiency and reducing storage requirements while maintaining accuracy comparable to the original large-scale model. MetaDistill \cite{zhou2022bert} offers a simple and efficient alternative to traditional KD methods by keeping the teacher model fixed during training. Within the meta-learning framework, teacher networks enhance knowledge transfer to student networks by distilling feedback on student performance. Additionally, a pilot update mechanism is introduced to improve the alignment between internal learners and meta-learners, focusing on enhancing internal learners' performance. Extensive experiments have validated the effectiveness and versatility of this method across text and image classification tasks. Furthermore, experiments on the GLUE benchmark have shown that MetaDistill significantly outperforms traditional knowledge distillation, achieving state-of-the-art performance compression. AD-KD \cite{wu-etal-2023-ad} addresses two key limitations of existing knowledge distillation methods. First, student models often merely mimic the teacher's behavior without developing their own reasoning capabilities. Second, these methods typically focus on transferring knowledge specific to complex models while neglecting data-specific knowledge. To overcome these issues, AD-KD introduces an innovative attribution-driven knowledge distillation method, which calculates the importance score of each input token using a gradient-based attribution approach \cite{5555}. To minimize the impact of less significant dimensions in the teacher's input embeddings, a top-K strategy filters out dimensions with lower attribution scores. The remaining scores are aggregated and normalized to reflect the importance of individual tokens. Additionally, this method extracts all potential predicted attribution knowledge, not just the highest probability prediction. To improve knowledge transfer for reasoning and generalization, AD-KD explores multi-view attribution distillation of all potential decisions made by the teacher. Experimental results on the GLUE benchmark indicate that this method surpasses several state-of-the-art approaches in performance.

Mukherjee \etal \cite{mukherjee} present XtremeDistil, a distillation method leveraging internal representations and parameter projections that are independent of the teacher's architecture. Unlike previous approaches focused on single-language GLUE tasks, this method distills multilingual Named Entity Recognition (NER) across 41 languages, using the multilingual bidirectional encoder representation from Transformers (mBERT) \cite{tsai} as the teacher model. Experimental results indicate that XtremeDistil achieves higher compression and faster inference speeds. Additionally, the study explored several previously unexamined aspects of distillation, including the effects of unlabeled transmission data and annotation resources, the selection of multilingual word embeddings, architectural modifications, and inference delays. This method significantly compressed the teacher model by up to 35 times in terms of parameters and reduced batch inference delay by 51 times while maintaining 95\% of the performance in large-scale multilingual NER and either matching or surpassing it in classification tasks. TinyBERT \cite{tinybert} integrates pre-trained distillation with fine-tuning distillation to capture both general domain and task-specific knowledge from BERT. It extracts multiple types of knowledge from different layers, including the embedding layer, hidden states, attention matrices, and transformation layers. During the GLUE benchmark evaluation, its teacher model BERT$_{base}$ achieved a performance exceeding 96.8\%, while offering inference speeds that were 7.5 to 9.4 times faster. MiniLM \cite{496209} introduced a depth self-attention distillation framework for task-agnostic Transformer-based language model (LM) distillation. This method isolates the self-attention module of the teacher model's final Transformer layer and uses the scaled dot-product between values within this module as a novel form of depth self-attention knowledge. This technique addresses the challenge of layer alignment between teacher and student models by transforming various dimensional representations of both models into a relation matrix of matching dimensionality, without requiring additional parameters for transforming student representations. This enhances the depth flexibility of the student model. MiniLM retained over 99\% accuracy on the SQuAD 2.0 \cite{Rajpurkar_Jia_Liang_2018} and various GLUE benchmark tasks while using only 50\% of the Transformer parameters and computational resources of the teacher model. This demonstrates the effectiveness of employing a teacher assistant \cite{mirzadeh2020improved} in distilling large pre-trained Transformer-based models. TED \cite{3619267} introduces an innovative task-aware layout distillation method designed to combat underfitting in student models and remove unnecessary information from teachers' hidden representations. This method aligns the hidden representations of students and teachers at each level, employing task-aware filters to extract relevant knowledge for the target task. By doing so, it narrows the knowledge gap between the models and enhances the student's ability to adapt to the target task. MobileBERT \cite{mobilebert} and HomoBERT \cite{liang2023homodistil} primarily focus on adjusting the model's width while maintaining its depth. This contrasts with Turc \etal \cite{turc2019well}, who found that altering model depth significantly impacts performance. MobileBERT introduces bottlenecks and inverted bottlenecks to both teacher and student models to modify hidden dimensions. However, this approach can disrupt the parameter balance between the multi-head attention and feed-forward networks, which is mitigated by using a stacked Feed-Forward Network (FFN) approach. Knowledge extraction is then carried out through the attention and hidden states of the transformer layers. HomoBERT, on the other hand, employs pruning. It starts by initializing the student model with the teacher model to ensure minimal initial divergence. It then targets input embeddings, hidden states, attention matrices, and output logits for pruning to create the distillation loss function. In each iteration, the most significant neurons are pruned based on importance scores, and the student model is trained using the distillation loss. This iterative process continues until the student model achieves the desired size. While white-box distillation is limited by the proprietary nature of LLMs, restricting its applicability, the rise of diverse open-source LLMs like Alpaca \cite{alpaca} and Vicuna \cite{chiang2023vicuna} offers promising prospects for the future of white-box distillation.

\subsection{Robustness Evaluation of White-box KD}

There are various evaluation standards for existing white-box KD algorithms, most of which utilize BERT as the teacher model. However, the effectiveness of these distillation algorithms in the context of LLMs remains unclear. Building on the work presented in \cite{wang2023robustness}, we conducted a unified evaluation of these algorithms from a robustness perspective, specifically focusing on adversarial robustness and out-of-distribution (OOD) robustness. Both types of robustness pertain to performance under input disturbances, which is particularly critical for safety-sensitive applications. Adversarial robustness examines the stability of models against adversarial and imperceptible disturbances, while OOD robustness assesses performance on unseen data that differs from the training data distribution. To evaluate adversarial robustness, we employed the AdvGLUE \cite{wang2021adversarial} and ANLI \cite{nie2020adversarial} benchmarks, using Attack Success Rate (ASR) as the metric. For OOD robustness, we used the Flipkart \cite{vaghani4940809flipkart} review and DDXPlus \cite{tchango2022ddxplus} medical diagnostic datasets, with F1-score (F1) as the indicator. Inspired by the work on MINILLM \cite{gu2023knowledge}, we utilized the Dolly $\footnote{https://github.com/databrickslabs/dolly/tree/master}$ dataset for distillation, fine-tuning both student and teacher models. We evaluated five distillation algorithms and four models concurrently to assess their robustness.

\begin{table}
\caption{The evaluation results of GPT-2. The optimal score for each model size is displayed in bold, and the score for student models outperforming teachers is marked with an $\ast$.}
\label{tab0000000}
\resizebox{\linewidth}{!}{
\begin{tabular}{@{}c|l|ccccccc|cc@{}}
\toprule
\multirow{2}{*}{Params} & \multicolumn{1}{c|}{\multirow{2}{*}{Method}} & \multicolumn{7}{c|}{Adversarial Robustness(ASR↓)}                                                                                                                                     & \multicolumn{2}{c}{OOD Robustness(F1↑)} \\ \cmidrule(l){3-11} 
                        & \multicolumn{1}{c|}{}                        & \multicolumn{1}{c|}{SST-2} & \multicolumn{1}{c|}{QQP}   & \multicolumn{1}{c|}{MNLI}  & \multicolumn{1}{c|}{QNLI}  & \multicolumn{1}{c|}{RTE}   & \multicolumn{1}{c|}{MNLI-MM} & ANLI  & \multicolumn{1}{c|}{Flipkart} & DDXPlus \\ \midrule
\multirow{1}{*}{1.5B}   & Teacher                                          & \multicolumn{1}{c|}{54.73} & \multicolumn{1}{c|}{96.15} & \multicolumn{1}{c|}{86.78} & \multicolumn{1}{c|}{93.92} & \multicolumn{1}{c|}{80.25} & \multicolumn{1}{c|}{87.65}   & 87.67 & \multicolumn{1}{c|}{38.55}    & 0.40    \\ \midrule
\multirow{6}{*}{120M}   & SFT                                          & \multicolumn{1}{c|}{57.43} & \multicolumn{1}{c|}{96.15} & \multicolumn{1}{c|}{91.74} & \multicolumn{1}{c|}{86.49} & \multicolumn{1}{c|}{61.73} & \multicolumn{1}{c|}{88.89}   & 96.17 & \multicolumn{1}{c|}{14.26}    & 0       \\
                        & KD                                           & \multicolumn{1}{c|}{83.78} & \multicolumn{1}{c|}{$\textbf{94.87}^{\ast}$} & \multicolumn{1}{c|}{89.26} & \multicolumn{1}{c|}{92.57} & \multicolumn{1}{c|}{69.14} & \multicolumn{1}{c|}{$\textbf{82.72}^{\ast}$}   & 94.08 & \multicolumn{1}{c|}{12.96}    & 0       \\
                        & SeqKD                                        & \multicolumn{1}{c|}{70.27} & \multicolumn{1}{c|}{96.15} & \multicolumn{1}{c|}{90.08} & \multicolumn{1}{c|}{87.84} & \multicolumn{1}{c|}{75.31} & \multicolumn{1}{c|}{83.95}   & \textbf{88.83} & \multicolumn{1}{c|}{12.61}    & 0       \\
                        & RKL                                          & \multicolumn{1}{c|}{66.89} & \multicolumn{1}{c|}{98.72} & \multicolumn{1}{c|}{\textbf{87.60}}  & \multicolumn{1}{c|}{85.81} & \multicolumn{1}{c|}{65.43} & \multicolumn{1}{c|}{93.83}   & 98.08 & \multicolumn{1}{c|}{7.46}     & 0       \\
                        & JS                                           & \multicolumn{1}{c|}{$\textbf{54.05}^{\ast}$} & \multicolumn{1}{c|}{100}   & \multicolumn{1}{c|}{94.21} & \multicolumn{1}{c|}{$\textbf{77.70}^{\ast}$} & \multicolumn{1}{c|}{74.07} & \multicolumn{1}{c|}{95.06}   & 97.83 & \multicolumn{1}{c|}{20.25}    & 0       \\
                        & MINILLM                                      & \multicolumn{1}{c|}{64.19} & \multicolumn{1}{c|}{100}   & \multicolumn{1}{c|}{89.26} & \multicolumn{1}{c|}{90.54} & \multicolumn{1}{c|}{$\textbf{53.09}^{\ast}$} & \multicolumn{1}{c|}{84.57}   & 95.50 & \multicolumn{1}{c|}{$\textbf{23.32}$}    & $\textbf{1.25}^{\ast}$    \\ \midrule
\multirow{6}{*}{340M}   & SFT                                          & \multicolumn{1}{c|}{70.27} & \multicolumn{1}{c|}{97.44} & \multicolumn{1}{c|}{92.56} & \multicolumn{1}{c|}{95.27} & \multicolumn{1}{c|}{$\textbf{55.56}^{\ast}$} & \multicolumn{1}{c|}{83.33}   & 94.00 & \multicolumn{1}{c|}{46.36}    & 0       \\
                        & KD                                           & \multicolumn{1}{c|}{63.51} & \multicolumn{1}{c|}{98.72} & \multicolumn{1}{c|}{84.30} & \multicolumn{1}{c|}{90.54} & \multicolumn{1}{c|}{72.84} & \multicolumn{1}{c|}{80.86}   & 92.42 & \multicolumn{1}{c|}{52.12}    & 0       \\
                        & SeqKD                                        & \multicolumn{1}{c|}{66.89} & \multicolumn{1}{c|}{97.44} & \multicolumn{1}{c|}{81.82} & \multicolumn{1}{c|}{93.24} & \multicolumn{1}{c|}{72.84} & \multicolumn{1}{c|}{79.01}   & 97.08 & \multicolumn{1}{c|}{47.87}    & 0       \\
                        & RKL                                          & \multicolumn{1}{c|}{62.16} & \multicolumn{1}{c|}{96.15} & \multicolumn{1}{c|}{95.04} & \multicolumn{1}{c|}{95.27} & \multicolumn{1}{c|}{70.37} & \multicolumn{1}{c|}{92.59}   & 96.92 & \multicolumn{1}{c|}{33.64}    & 0       \\
                        & JS                                           & \multicolumn{1}{c|}{62.16} & \multicolumn{1}{c|}{96.15} & \multicolumn{1}{c|}{95.04} & \multicolumn{1}{c|}{95.27} & \multicolumn{1}{c|}{70.37} & \multicolumn{1}{c|}{92.59}   & 96.92 & \multicolumn{1}{c|}{32.29}    & 0       \\
                        & MINILLM                                      & \multicolumn{1}{c|}{$\textbf{54.73}^{\ast}$} & \multicolumn{1}{c|}{$\textbf{92.31}^{\ast}$} & \multicolumn{1}{c|}{$\textbf{78.51}^{\ast}$} & \multicolumn{1}{c|}{$\textbf{87.16}^{\ast}$} & \multicolumn{1}{c|}{69.14} & \multicolumn{1}{c|}{$\textbf{75.31}^{\ast}$}   & \textbf{88.75} & \multicolumn{1}{c|}{$\textbf{52.45}^{\ast}$}    & $\textbf{0.68}^{\ast}$    \\ \midrule
\multirow{6}{*}{760M}   & SFT                                          & \multicolumn{1}{c|}{56.76} & \multicolumn{1}{c|}{97.44} & \multicolumn{1}{c|}{90.91} & \multicolumn{1}{c|}{92.57} & \multicolumn{1}{c|}{80.25} & \multicolumn{1}{c|}{91.36}   & 91.75 & \multicolumn{1}{c|}{28.52}    & 0       \\
                        & KD                                           & \multicolumn{1}{c|}{$\textbf{53.38}^{\ast}$} & \multicolumn{1}{c|}{$\textbf{94.87}^{\ast}$} & \multicolumn{1}{c|}{90.91} & \multicolumn{1}{c|}{95.27} & \multicolumn{1}{c|}{64.20} & \multicolumn{1}{c|}{87.65}   & 95.58 & \multicolumn{1}{c|}{30.77}    & 0       \\
                        & SeqKD                                        & \multicolumn{1}{c|}{55.41} & \multicolumn{1}{c|}{96.15} & \multicolumn{1}{c|}{90.91} & \multicolumn{1}{c|}{97.30} & \multicolumn{1}{c|}{80.25} & \multicolumn{1}{c|}{89.51}   & 94.17 & \multicolumn{1}{c|}{27.60}    & 0       \\
                        & RKL                                          & \multicolumn{1}{c|}{64.86} & \multicolumn{1}{c|}{97.44} & \multicolumn{1}{c|}{90.91} & \multicolumn{1}{c|}{92.57} & \multicolumn{1}{c|}{85.19} & \multicolumn{1}{c|}{95.06}   & 98.67 & \multicolumn{1}{c|}{20.53}    & 0       \\
                        & JS                                           & \multicolumn{1}{c|}{60.14} & \multicolumn{1}{c|}{98.72} & \multicolumn{1}{c|}{95.04} & \multicolumn{1}{c|}{94.59} & \multicolumn{1}{c|}{77.78} & \multicolumn{1}{c|}{96.30}   & 98.58 & \multicolumn{1}{c|}{19.90}    & 0       \\
                       & MINILLM                                      & \multicolumn{1}{c|}{54.05} & \multicolumn{1}{c|}{96.15} & \multicolumn{1}{c|}{$\textbf{81.82}^{\ast}$} & \multicolumn{1}{c|}{$\textbf{89.19}^{\ast}$} & \multicolumn{1}{c|}{$\textbf{60.49}^{\ast}$} & \multicolumn{1}{c|}{$\textbf{72.22}^{\ast}$}   & $\textbf{85.33}^{\ast}$ & \multicolumn{1}{c|}{$\textbf{47.00}^{\ast}$}    & 0       \\ \bottomrule
\end{tabular}
}
\end{table}

\begin{table}
\caption{The evaluation results of OPT.}
\label{tab00}
\resizebox{\linewidth}{!}{
\begin{tabular}{@{}c|l|ccccccc|cc@{}}
\toprule
                         & \multicolumn{1}{c|}{}                         & \multicolumn{7}{c|}{{\color[HTML]{000000} Adversarial Robustness(ASR↓)}}                                                                                                                                                                                                                                                                        & \multicolumn{2}{c}{{\color[HTML]{000000} OOD Robustness(F1↑)}}                        \\ \cmidrule(l){3-11} 
\multirow{-2}{*}{Params} & \multicolumn{1}{c|}{\multirow{-2}{*}{Method}} & \multicolumn{1}{c|}{{\color[HTML]{000000} SST-2}} & \multicolumn{1}{c|}{{\color[HTML]{000000} QQP}} & \multicolumn{1}{c|}{{\color[HTML]{000000} MNLI}} & \multicolumn{1}{c|}{{\color[HTML]{000000} QNLI}} & \multicolumn{1}{c|}{{\color[HTML]{000000} RTE}} & \multicolumn{1}{c|}{{\color[HTML]{000000} MNLI-MM}} & {\color[HTML]{000000} ANLI} & \multicolumn{1}{c|}{{\color[HTML]{000000} Flipkart}} & {\color[HTML]{000000} DDXPlus} \\ \midrule
13B                     & Teacher                                       & \multicolumn{1}{c|}{52.70}                        & \multicolumn{1}{c|}{94.87}                      & \multicolumn{1}{c|}{76.03}                       & \multicolumn{1}{c|}{91.89}                       & \multicolumn{1}{c|}{61.73}                      & \multicolumn{1}{c|}{70.99}                          & 99.58                       & \multicolumn{1}{c|}{54.81}                           & 0.18                           \\ \midrule
                         & SFT                                           & \multicolumn{1}{c|}{54.05}                        & \multicolumn{1}{c|}{94.87}                      & \multicolumn{1}{c|}{78.51}                       & \multicolumn{1}{c|}{94.59}                       & \multicolumn{1}{c|}{61.73}                      & \multicolumn{1}{c|}{77.78}                          & 98.00                       & \multicolumn{1}{c|}{51.06}                           & $\textbf{1.84}^{\ast}$                           \\
                         & KD                                            & \multicolumn{1}{c|}{52.03}                        & \multicolumn{1}{c|}{$\textbf{65.38}^{\ast}$}                      & \multicolumn{1}{c|}{\textbf{77.69}}                       & \multicolumn{1}{c|}{$\textbf{68.92}^{\ast}$}                       & \multicolumn{1}{c|}{50.62}                      & \multicolumn{1}{c|}{$\textbf{69.14}^{\ast}$}                          & 87.67                       & \multicolumn{1}{c|}{44.45}                           & 0                              \\
                         & SeqKD                                         & \multicolumn{1}{c|}{48.65}                        & \multicolumn{1}{c|}{93.59}                      & \multicolumn{1}{c|}{80.17}                       & \multicolumn{1}{c|}{85.14}                       & \multicolumn{1}{c|}{83.95}                      & \multicolumn{1}{c|}{75.31}                          & $\textbf{78.17}^{\ast}$                       & \multicolumn{1}{c|}{28.85}                           & 0.30                           \\
                         & RKL                                           & \multicolumn{1}{c|}{$\textbf{45.95}^{\ast}$}                        & \multicolumn{1}{c|}{94.87}                      & \multicolumn{1}{c|}{92.56}                       & \multicolumn{1}{c|}{97.30}                       & \multicolumn{1}{c|}{74.07}                      & \multicolumn{1}{c|}{90.12}                          & 99.00                       & \multicolumn{1}{c|}{46.41}                           & 0                              \\
                         & JS                                            & \multicolumn{1}{c|}{47.97}                        & \multicolumn{1}{c|}{82.05}                      & \multicolumn{1}{c|}{80.99}                       & \multicolumn{1}{c|}{89.86}                       & \multicolumn{1}{c|}{48.15}                      & \multicolumn{1}{c|}{89.51}                          & 94.67                       & \multicolumn{1}{c|}{45.15}                           & 0                              \\
\multirow{-6}{*}{1.3B}   & MINILLM                                       & \multicolumn{1}{c|}{47.30}                        & \multicolumn{1}{c|}{94.87}                      & \multicolumn{1}{c|}{78.51}                       & \multicolumn{1}{c|}{79.73}                       & \multicolumn{1}{c|}{$\textbf{49.38}^{\ast}$}                      & \multicolumn{1}{c|}{71.60}                          & 89.75                       & \multicolumn{1}{c|}{$\textbf{55.88}^{\ast}$}                           & 0.13                           \\ \midrule
                         & SFT                                           & \multicolumn{1}{c|}{51.35}                        & \multicolumn{1}{c|}{96.15}                      & \multicolumn{1}{c|}{92.56}                       & \multicolumn{1}{c|}{97.30}                       & \multicolumn{1}{c|}{69.14}                      & \multicolumn{1}{c|}{93.83}                          & 98.17                       & \multicolumn{1}{c|}{47.86}                           & 0                              \\
                         & KD                                            & \multicolumn{1}{c|}{$\textbf{44.59}^{\ast}$}                        & \multicolumn{1}{c|}{84.62}                      & \multicolumn{1}{c|}{$\textbf{73.55}^{\ast}$}                       & \multicolumn{1}{c|}{83.78}                       & \multicolumn{1}{c|}{64.20}                      & \multicolumn{1}{c|}{$\textbf{66.05}^{\ast}$}                          & $\textbf{70.67}^{\ast}$                       & \multicolumn{1}{c|}{41.83}                           & 0                              \\
                         & SeqKD                                         & \multicolumn{1}{c|}{47.97}                        & \multicolumn{1}{c|}{$\textbf{75.64}^{\ast}$}                      & \multicolumn{1}{c|}{77.69}                       & \multicolumn{1}{c|}{$\textbf{77.03}^{\ast}$}                       & \multicolumn{1}{c|}{$\textbf{61.73}$}                      & \multicolumn{1}{c|}{72.22}                          & 88.50                       & \multicolumn{1}{c|}{34.85}                           & 0                              \\
                         & RKL                                           & \multicolumn{1}{c|}{56.08}                        & \multicolumn{1}{c|}{98.72}                      & \multicolumn{1}{c|}{90.91}                       & \multicolumn{1}{c|}{99.32}                       & \multicolumn{1}{c|}{100}                        & \multicolumn{1}{c|}{95.68}                          & 96.83                       & \multicolumn{1}{c|}{53.01}                           & $\textbf{0.29}^{\ast}$                          \\
                         & JS                                            & \multicolumn{1}{c|}{67.57}                        & \multicolumn{1}{c|}{96.15}                      & \multicolumn{1}{c|}{86.78}                       & \multicolumn{1}{c|}{97.97}                       & \multicolumn{1}{c|}{75.31}                      & \multicolumn{1}{c|}{89.51}                          & 92.25                       & \multicolumn{1}{c|}{46.55}                           & 0                              \\
\multirow{-6}{*}{2.7B}   & MINILLM                                       & \multicolumn{1}{c|}{47.30}                        & \multicolumn{1}{c|}{94.87}                      & \multicolumn{1}{c|}{$\textbf{73.55}^{\ast}$}                       & \multicolumn{1}{c|}{91.89}                       & \multicolumn{1}{c|}{66.67}                      & \multicolumn{1}{c|}{72.84}                          & 81.17                       & \multicolumn{1}{c|}{$\textbf{58.09}^{\ast}$}                           & 0.09                           \\ \midrule
                         & SFT                                           & \multicolumn{1}{c|}{49.32}                        & \multicolumn{1}{c|}{89.74}                      & \multicolumn{1}{c|}{85.12}                       & \multicolumn{1}{c|}{81.76}                       & \multicolumn{1}{c|}{77.78}                      & \multicolumn{1}{c|}{79.63}                          & $\textbf{75.58}^{\ast}$                      & \multicolumn{1}{c|}{43.05}                           & 0.06                           \\
                         & KD                                            & \multicolumn{1}{c|}{$\textbf{45.27}^{\ast}$}                        & \multicolumn{1}{c|}{94.87}                      & \multicolumn{1}{c|}{80.17}                       & \multicolumn{1}{c|}{85.14}                       & \multicolumn{1}{c|}{58.02}                      & \multicolumn{1}{c|}{\textbf{73.46}}                          & 81.33                       & \multicolumn{1}{c|}{50.99}                           & 0                              \\
                         & SeqKD                                         & \multicolumn{1}{c|}{50.68}                        & \multicolumn{1}{c|}{92.31}                      & \multicolumn{1}{c|}{85.95}                       & \multicolumn{1}{c|}{77.70}                       & \multicolumn{1}{c|}{86.42}                      & \multicolumn{1}{c|}{80.86}                          & 78.17                       & \multicolumn{1}{c|}{32.01}                           & $\textbf{0.65}^{\ast}$                           \\
                         & RKL                                           & \multicolumn{1}{c|}{60.81}                        & \multicolumn{1}{c|}{91.03}                      & \multicolumn{1}{c|}{95.87}                       & \multicolumn{1}{c|}{88.51}                       & \multicolumn{1}{c|}{$\textbf{49.38}^{\ast}$}                      & \multicolumn{1}{c|}{93.21}                          & 97.92                       & \multicolumn{1}{c|}{24.16}                           & 0                              \\
                         & JS                                            & \multicolumn{1}{c|}{63.51}                        & \multicolumn{1}{c|}{$\textbf{88.46}^{\ast}$}                      & \multicolumn{1}{c|}{94.21}                       & \multicolumn{1}{c|}{$\textbf{69.59}^{\ast}$}                       & \multicolumn{1}{c|}{62.96}                      & \multicolumn{1}{c|}{88.27}                          & 98.58                       & \multicolumn{1}{c|}{26.05}                           & 0                              \\
\multirow{-6}{*}{6.7B}   & MINILLM                                       & \multicolumn{1}{c|}{50.68}                        & \multicolumn{1}{c|}{92.31}                      & \multicolumn{1}{c|}{$\textbf{74.38}^{\ast}$}                       & \multicolumn{1}{c|}{95.27}                       & \multicolumn{1}{c|}{80.25}                      & \multicolumn{1}{c|}{75.93}                          & 86.75                       & \multicolumn{1}{c|}{$\textbf{56.43}^{\ast}$}                           & 0                              \\ \bottomrule
\end{tabular}
}
\end{table}

\begin{table}
\caption{The evaluation results of LLaMA and LLaMA2.}
\label{tab000}
\resizebox{\linewidth}{!}{
\begin{tabular}{@{}c|l|ccccccc|cc@{}}
\toprule
                         & \multicolumn{1}{c|}{}                         & \multicolumn{7}{c|}{{\color[HTML]{000000} Adversarial Robustness(ASR↓)}}                                                                                                                                                                                                                                                                        & \multicolumn{2}{c}{{\color[HTML]{000000} OOD Robustness(F1↑)}}                        \\ \cmidrule(l){3-11} 
\multirow{-2}{*}{Params} & \multicolumn{1}{c|}{\multirow{-2}{*}{Method}} & \multicolumn{1}{c|}{{\color[HTML]{000000} SST-2}} & \multicolumn{1}{c|}{{\color[HTML]{000000} QQP}} & \multicolumn{1}{c|}{{\color[HTML]{000000} MNLI}} & \multicolumn{1}{c|}{{\color[HTML]{000000} QNLI}} & \multicolumn{1}{c|}{{\color[HTML]{000000} RTE}} & \multicolumn{1}{c|}{{\color[HTML]{000000} MNLI-MM}} & {\color[HTML]{000000} ANLI} & \multicolumn{1}{c|}{{\color[HTML]{000000} Flipkart}} & {\color[HTML]{000000} DDXPlus} \\ \midrule
13B                      & Teacher                                       & \multicolumn{1}{c|}{41.89}                        & \multicolumn{1}{c|}{65.38}                      & \multicolumn{1}{c|}{76.03}                       & \multicolumn{1}{c|}{52.70}                       & \multicolumn{1}{c|}{56.79}                      & \multicolumn{1}{c|}{64.20}                          & 66.42                       & \multicolumn{1}{c|}{56.01}                           & 9.09                           \\ \midrule
                         & SFT                                           & \multicolumn{1}{c|}{47.97}                        & \multicolumn{1}{c|}{67.95}                      & \multicolumn{1}{c|}{71.90}                       & \multicolumn{1}{c|}{48.65}                       & \multicolumn{1}{c|}{46.91}                      & \multicolumn{1}{c|}{66.67}                          & 67.33                       & \multicolumn{1}{c|}{52.65}                           & 2.81                           \\
                         & KD                                            & \multicolumn{1}{c|}{43.24}                        & \multicolumn{1}{c|}{60.26}                      & \multicolumn{1}{c|}{73.55}                       & \multicolumn{1}{c|}{54.73}                       & \multicolumn{1}{c|}{56.79}                      & \multicolumn{1}{c|}{\textbf{64.20}}                          & \textbf{66.58}                       & \multicolumn{1}{c|}{$\textbf{65.16}^{\ast}$}                           & 4.32                           \\
                         & SeqKD                                         & \multicolumn{1}{c|}{$\textbf{40.54}^{\ast}$}                        & \multicolumn{1}{c|}{65.38}                      & \multicolumn{1}{c|}{$\textbf{67.77}^{\ast}$}                       & \multicolumn{1}{c|}{49.32}                       & \multicolumn{1}{c|}{$\textbf{41.98}^{\ast}$}                      & \multicolumn{1}{c|}{64.81}                          & 66.67                       & \multicolumn{1}{c|}{56.58}                           & \textbf{5.66}                           \\
                         & RKL                                           & \multicolumn{1}{c|}{$\textbf{40.54}^{\ast}$}                        & \multicolumn{1}{c|}{76.92}                      & \multicolumn{1}{c|}{72.73}                       & \multicolumn{1}{c|}{$\textbf{45.27}^{\ast}$}                       & \multicolumn{1}{c|}{50.62}                      & \multicolumn{1}{c|}{64.81}                          & 67.75                       & \multicolumn{1}{c|}{52.74}                           & 0.43                           \\
                         & JS                                            & \multicolumn{1}{c|}{43.92}                        & \multicolumn{1}{c|}{66.67}                      & \multicolumn{1}{c|}{73.55}                       & \multicolumn{1}{c|}{52.70}                       & \multicolumn{1}{c|}{56.79}                      & \multicolumn{1}{c|}{\textbf{64.20}}                          & 67.00                       & \multicolumn{1}{c|}{57.35}                           & 3.44                           \\
\multirow{-6}{*}{LLaMA\_7B}     & MINILLM                                       & \multicolumn{1}{c|}{42.57}                        & \multicolumn{1}{c|}{$\textbf{48.72}^{\ast}$}                      & \multicolumn{1}{c|}{71.90}                       & \multicolumn{1}{c|}{51.35}                       & \multicolumn{1}{c|}{56.79}                      & \multicolumn{1}{c|}{64.81}                          & 66.92                       & \multicolumn{1}{c|}{52.27}                           & 5.21                           \\ \midrule
13B                      & Teacher                                       & \multicolumn{1}{c|}{54.05}                        & \multicolumn{1}{c|}{47.44}                      & \multicolumn{1}{c|}{65.29}                       & \multicolumn{1}{c|}{51.35}                       & \multicolumn{1}{c|}{46.91}                      & \multicolumn{1}{c|}{62.96}                          & 71.00                       & \multicolumn{1}{c|}{29.76}                           & 0                              \\ \midrule
                         & SFT                                           & \multicolumn{1}{c|}{$\textbf{43.92}^{\ast}$}                        & \multicolumn{1}{c|}{69.23}                      & \multicolumn{1}{c|}{75.21}                       & \multicolumn{1}{c|}{63.51}                       & \multicolumn{1}{c|}{61.73}                      & \multicolumn{1}{c|}{66.05}                          & 69.92                       & \multicolumn{1}{c|}{50.76}                           & 0.17                           \\
                         & KD                                            & \multicolumn{1}{c|}{47.97}                        & \multicolumn{1}{c|}{70.51}                      & \multicolumn{1}{c|}{74.38}                       & \multicolumn{1}{c|}{52.70}                       & \multicolumn{1}{c|}{56.79}                      & \multicolumn{1}{c|}{64.20}                          & 83.67                       & \multicolumn{1}{c|}{55.51}                           & 1.75                           \\
                         & SeqKD                                         & \multicolumn{1}{c|}{49.32}                        & \multicolumn{1}{c|}{66.67}                      & \multicolumn{1}{c|}{75.21}                       & \multicolumn{1}{c|}{52.70}                       & \multicolumn{1}{c|}{56.79}                      & \multicolumn{1}{c|}{63.59}                          & 84.50                       & \multicolumn{1}{c|}{$\textbf{55.63}^{\ast}$}                           & 1.67                           \\
                         & RKL                                           & \multicolumn{1}{c|}{52.03}                        & \multicolumn{1}{c|}{61.54}                      & \multicolumn{1}{c|}{\textbf{73.55}}                       & \multicolumn{1}{c|}{$\textbf{47.97}^{\ast}$}                       & \multicolumn{1}{c|}{58.02}                      & \multicolumn{1}{c|}{\textbf{63.58}}                          & 72.33                       & \multicolumn{1}{c|}{55.42}                           & 4.37                           \\
                         & JS                                            & \multicolumn{1}{c|}{51.35}                        & \multicolumn{1}{c|}{$\textbf{57.69}$}                      & \multicolumn{1}{c|}{78.51}                       & \multicolumn{1}{c|}{56.76}                       & \multicolumn{1}{c|}{$\textbf{37.04}^{\ast}$}                      & \multicolumn{1}{c|}{66.67}                          & $\textbf{69.00}^{\ast}$                      & \multicolumn{1}{c|}{50.38}                           & $\textbf{12.19}^{\ast}$                          \\
\multirow{-6}{*}{LLaMA2\_7B}     & MINILLM                                       & \multicolumn{1}{c|}{50.00}                        & \multicolumn{1}{c|}{75.64}                      & \multicolumn{1}{c|}{75.21}                       & \multicolumn{1}{c|}{58.78}                       & \multicolumn{1}{c|}{48.15}                      & \multicolumn{1}{c|}{69.75}                          & 83.67                       & \multicolumn{1}{c|}{31.94}                           & 0                              \\ \bottomrule
\end{tabular}
}
\end{table}

The evaluation results are shown in Tables \ref{tab0000000}-\ref{tab000}. Firstly, we observed that MINILLM demonstrated superior overall distillation performance in GPT-2. Notably, for the 340M-sized GPT-2, it achieved state-of-the-art results on both adversarial and out-of-distribution datasets when compared to the other four distillation algorithms. Furthermore, MINILLM outperformed the other algorithms on the Flipkart and DDXPlus datasets for GPT-2 of any size, highlighting its exceptional generalization capability to out-of-distribution data. Secondly, for the OPT model, we discovered that the most straightforward KD algorithm, which employs the teacher distribution as supervision for each token step to fine-tune the student model, achieved the best overall performance. Likewise, MINILLM outperformed other distillation algorithms and even exceeded the performance of teacher models for OPTs of any size on the Flipkart dataset. Finally, for LLaMA, SeqKD demonstrated a comparatively better distillation effect, whereas for LLaMA2, JS showed a relatively superior performance. This suggests that even when the model size is identical and the model structure is similar, the effectiveness of the same distillation algorithm can vary significantly.

\subsection{Discussion on White-box KD}

Logits-based KD methods typically focus on aligning the output distributions between the teacher and student models. In contrast, hint-based KD methods can convey richer information by aligning the intermediate layers, leading to better results. However, implementing layer-to-layer knowledge distillation necessitates careful design of the layer mappings between the teacher and student models and requires a deep understanding of the model architecture. Both logits-based and hint-based KD methods demand substantial GPU memory during the distillation process. Even though the teacher network doesn't need backpropagation, the activation of intermediate features during forward propagation consumes a significant amount of GPU memory. Therefore, exploring ways to reduce training costs and shorten training times is crucial.

\subsection{Black-box Knowledge Distillation}

The two previously discussed distillation techniques rely on access to the internal data of the teacher model, categorizing them as white-box distillation methods, which require internal data during training. However, many modern large-scale closed-source models do not provide access to internal data, limiting us to using only model predictions. Distillation where knowledge is transferred solely through the teacher model's predictions is known as black-box knowledge distillation. Researchers have found that when model parameters are sufficiently large, the models exhibit remarkable versatility, enabling them to handle complex tasks. Many black-box distillation methods take advantage of this capability, typically utilizing three techniques: In-Context Learning, Chain-of-Thought, and Instruction Following. In this section, we further categorize black-box KD methods based on the use of emergent capabilities.

\begin{table}
\caption{Comparison of representative black-box KD methods on large language models.}
\label{tab3}
\resizebox{\linewidth}{!}{
\begin{tabular}{@{}llcccc@{}}
\toprule
Models & Distillation Type  & Teacher Model  & Compression Rate  & Evaluation Task & Comparison with Teacher Model

\\ \midrule
ILD \cite{huang2022context}    & ICL  &  \begin{tabular}[c]{@{}l@{}}BERT$_{large}$ \\ GPT-2$_{large}$ \end{tabular} & \begin{tabular}[c]{@{}l@{}}13$\times$ \\ 6$\times$ \end{tabular} & \begin{tabular}[c]{@{}l@{}}  LAMA\cite{petroni2019language}\\  CrossFit\cite{ye2021crossfit} \end{tabular} &  \begin{tabular}[c]{@{}l@{}}  52.4/57.3 (91\% performance)  \\ 61.2/66.2 (92\% performance) \end{tabular} \\   \midrule
LLM-R \cite{wang2023learning}    & ICL  &  LLaMA$_{13B}$  & 2$\times$  &  Commonsense/Coreference/NLI/Paraphrase/Sentiment/Data-to-text/Summarize/etc.(30)  &  68.8/64.6 (107\% performance)   \\ \midrule
MT-CoT \cite{liang2020mixkd}    & CoT  &  GPT-3$_{text-davinci-002}$  & 58$\times$   &  CommonsenseQA\cite{talmor2019commonsenseqa}/StrategyQA\cite{geva2021did}/OpenbookQA\cite{mihaylov2018can}  &  80.5/82.1 (98\% performance) \\
Distilling step-by-step \cite{hsieh2023distilling}    & CoT  &  PaLM$_{540B}$ & 2455$\times$   & e-SNLI\cite{camburu2018snli}/ANLI\cite{nie2020adversarial}/CQA\cite{talmor2019commonsenseqa}/SVAMP\cite{patel2021nlp} &  58.4/72.3 (81\% performance)  \\ 
Fine-tune-CoT \cite{ho2022large} & CoT  &  InstructGPT 175B$_{text-davinci-002}$  & 26$\times$     &  SingleEq/AddSub/MultiArith/GSM8K/AQUA-RAT/SVAMP/StrategyQA/etc.(12)   &  42.2/65.5 (64\% performance) \\
MCC-KD \cite{chen2023mcc}   & CoT  &  GPT-3.5$_{Turbo}$  & /     &  GSM8K\cite{cobbe2021training}/ASDiv\cite{miao2020diverse}/SVAMP\cite{patel2021nlp}/CommonsenseQA\cite{talmor2019commonsenseqa}   &  66.2/75.8 (87\% performance) \\

SCOTT \cite{wang2023scott} & CoT  &  GPT-neox$_{20B}$  & 7$\times$     &  CSQA\cite{talmor2019commonsenseqa}/CREAK\cite{onoe2021creak}/QASC\cite{khot2020qasc}/StrategyQA\cite{geva2021did}  
&  69.6/72.3 (96\% performance) \\
LTD \cite{jie2023leveraging}  & CoT  &  Codex$_{code-davinci-002}$  & 29$\times$     &  SVAMP\cite{patel2021nlp}/GSM8K\cite{cobbe2021training}/MathQA\cite{amini2019mathqa} 
&  45.9/56.8 (81\% performance) \\ 
PaD \cite{zhu2023pad}    & CoT  &  PaLM$_{60B}$  & 78$\times$     &  SVAMP\cite{patel2021nlp}/GSM8K\cite{cobbe2021training}/ASDiv\cite{miao2020diverse}/MultiArith\cite{roy2015solving}/BBH\cite{srivastava2023beyond}
&  43.9/50.2 (88\% performance) \\  
Learn-to-Reason \cite{chae2023dialogue}    & CoT  &  ChatGPT$_{175B}$  & 29$\times$     &  GSM8K\cite{cobbe2021training}/MultiArith\cite{roy2015solving}/SVAMP\cite{patel2021nlp}/CSQA\cite{talmor2019commonsenseqa}/StrategyQA\cite{geva2021did}
&  62.1/76.1 (82\% performance) \\  \midrule
LaMini-LM \cite{wu2023lamini}    & IF  &  Alpaca$_{7B}$  & 9$\times$     &  Multiple-Choice QA/Extractive QA/Sentiment Analysis/Paraphrase Identification/etc.(15)
&  60.8/62.3 (98\% performance) \\ 
Lion \cite{jiang2023lion}    & IF  &  ChatGPT$_{175B}$  & 25$\times$     &  BBH\cite{srivastava2023beyond}
&  32.0/48.9 (65\% performance) \\  
UniNER \cite{zhou2023universalner}    & IF  &  GPT-3.5$_{turbo-0301}$  & /     &  UNIVERSAL NER BENCHMARK \cite{zhou2023universalner}
&  41.7/34.9 (119\% performance) \\ 

\bottomrule
\end{tabular}
}
\end{table}

\subsubsection{In-Context Learning}

ICL was initially introduced in GPT-3 \cite{brown2020language}, where it employs a natural language prompt that includes both task descriptions and multiple task examples as demonstrations. The process begins with the task description, followed by selecting specific instances from the task dataset to serve as examples. These instances are then formatted into natural language prompts using a predefined template and arranged in a particular order. Finally, the test samples are incorporated into the input of the LLM to produce the output.

Expanding on this concept, Huang et al. \cite{huang2022context} propose In-Context Learning Distillation, which aims to enhance the few-shot learning capabilities of multitask models by effectively extracting and transferring knowledge through context learning and language modeling objectives. This approach introduces two paradigms for few-shot learning: Meta In-context Tuning and Multitask In-context Tuning. In Meta-ICT \cite{chen2022meta, min2022metaicl}, the language model undergoes meta-training across a broad spectrum of tasks using in-context learning objectives. Subsequently, it adapts to unseen target tasks through in-context learning. However, the efficacy of in-context learning heavily relies on the knowledge accumulated during pretraining \cite{reynolds2021prompt}, potentially limiting its ability to fully leverage the input-label correspondence provided in the training data \cite{workrethinking}. To address this limitation, an alternative few-shot learning paradigm called Multitask In-Context Tuning is proposed. While Meta-ICT enables the student model to adapt to new tasks via context learning and teacher guidance, Multitask-ICT treats all target tasks as training tasks and utilizes examples directly from these tasks for in-context learning distillation. These two paradigms for few-shot learning involve a trade-off between performance and computational efficiency. Results across tasks such as classification, natural language inference, and question answering indicate that Multitask-ICT achieves a reduction in model size by 93\% while retaining 91.4\% of the teacher's performance. Therefore, Multitask-ICT proves to be more effective, albeit with higher computational costs. LLM-R \cite{wang2023learning} utilizes a pre-trained frozen LLM to retrieve high-quality contextual examples, which are then ranked to generate training data. Subsequently, it constructs a reward model using a cross-encoder to capture ranking preferences. Finally, knowledge distillation is applied to train a dense retriever based on dual encoders. Our comprehensive evaluation of LLM-R across diverse tasks consistently demonstrates superior performance compared to several robust baselines. Furthermore, our model exhibits scalability across different task sizes and LLM architectures. Detailed analysis indicates that our approach enhances context learning performance by an average of 7.8\%, with consistent improvements observed across various sizes of LLMs.

\subsubsection{Chain-of-Thought}

Chain-of-Thought (CoT) \cite{li2022explanations, magister2022teaching, hsieh2023distilling, wadhwa2023revisiting} represents an advanced prompting strategy aimed at enhancing LLMs' ability to tackle complex reasoning tasks. Unlike the input-output pair approach used in ICL for prompt formulation, CoT integrates intermediate inference steps that incorporate final outputs into the prompts. Typically, CoT distillation \cite{li2022explanations, ho2022large, shridhar2023distilling, wang2023scott, kang2024knowledge, jie2023leveraging, zhu2023pad, li2023symbolic, chen2023mcc, chae2023dialogue, wang2023democratizing} involves leveraging large-scale models to construct enriched datasets focused on reasoning tasks, which are then utilized for fine-tuning student models. Thus, the primary focus is on generating high-quality rationales for training and ensuring effective utilization of these rationales by students \cite{li2022explanations, hsieh2023distilling, shridhar2023distilling, wang2023scott, kang2024knowledge}.

Li \etal \cite{li2022explanations} pioneered the use of explanations generated by LLMs to enhance the training of smaller inference machines. They systematically explored three methods for deriving interpretations from LLMs and integrated them into a multitask learning framework to empower compact models with robust reasoning and interpretative capabilities.  Across multiple inference tasks, experiments consistently demonstrated that their approach outperforms baseline fine-tuning methods under various conditions. Notably, it achieved up to a 9.5\% accuracy improvement over GPT-3 (175B) after 60 rounds of fine-tuning on Commonsense QA. The high-quality explanations generated by their method elucidate the rationale behind AI's interpretable predictions. Hsieh \etal \cite{hsieh2023distilling} introduced step-by-step distilling, a novel and straightforward approach aimed at reducing the amount of training data required to refine and fine-tune LLMs into smaller models. Central to their method is a paradigm shift: LLMs are not merely sources of noisy labels but proxies capable of providing natural language reasoning to justify their predictions. Empirical findings across four NLP benchmark tests yielded three notable outcomes. Firstly, compared to fine-tuning and traditional distillation methods, their model reduced the average number of training samples required by over 50\% (with some reductions exceeding 85\%), leading to improved performance. Secondly, their model achieved superior performance to LLMs while being significantly smaller in size, thereby reducing computational resources for deployment. Thirdly, their method concurrently reduced model size and required data to outperform LLMs. For example, their final iteration of the 770M T5 model surpassed the performance of a 540B parameter LLM, utilizing only 80\% of the labeled dataset.

Moreover, Ho \etal \cite{ho2022large} propose fine-tuning CoT, a method harnessing LLMs' reasoning capabilities to guide smaller models in solving complex tasks. By generating multiple inference solutions from the teacher model through random sampling, they enrich the training data of the student model. Evaluation across 12 tasks using widely accessible models demonstrates that fine-tuning CoT achieves significant inference performance in smaller models while preserving much of the generality of hint-based CoT inference, previously reliant on models with over 100 billion parameters. Consequently, models with as few as 0.3 billion parameters can outperform larger counterparts in specific tasks, even surpassing the performance of the teacher model with 175 billion parameters. Similarly, Chen \etal \cite{chen2023mcc} introduced Multi-CoT Consistent Knowledge Distillation (MCC-KD) to efficiently capture the diversity and coherence of reasoning capabilities. In MCC-KD, multiple fundamental principles are generated for each question, and the consistency between corresponding predictions is strengthened by minimizing bidirectional KL divergence between answer distributions. MCC-KD's efficacy is evaluated on mathematical reasoning and common sense reasoning benchmarks across various model architectures. Empirical findings not only confirm MCC-KD's superior performance on in-distribution datasets but also highlight its robust generalization ability on out-of-distribution datasets. Fu \etal \cite{fu2023specializing} Fu \etal \cite{fu2023specializing} apply CoT to specialize smaller language models for multi-step mathematical reasoning tasks. The SOCRATIC CoT method, as detailed by Shridhar \etal \cite{shridhar2023distilling}, decomposes the original problem into a series of sub-problems and employs a pair of compact distillation models: a problem decomposer and a sub-problem solver. These models collaborate to break down and resolve complex problems presented in new tasks. Evaluation across various inference datasets, including GSM8K, StrategyQA, and SVAMP, demonstrates that this distillation approach significantly enhances the performance of smaller models by over 70\% compared to the baseline. On the other hand, SCOTT \cite{wang2023scott} introduces a core principle of leveraging a LLM to guide the correct answer through comparative decoding. This method encourages the teacher model to generate tokens that align closely with the correct answer, thereby improving the fidelity of the distillation process. Jie \etal \cite{jie2023leveraging} and Zhu \etal \cite{zhu2023pad} enhance mathematical reasoning capabilities through program distillation. Chae \etal \cite{chae2023dialogue} and Wang \etal \cite{wang2023democratizing} propose an interactive multi-loop learning framework. In this framework, the former focuses on training students using multi-hop reasoning, while the latter actively communicates their learning status to the LLM teacher. Subsequently, the teacher offers customized explanations for the students' feedback, guiding them to reflect on their errors.

\subsubsection{Instruction Following}

The instruction following capability aims to enhance the language model's ability to perform new tasks without heavy reliance on limited examples. Through fine-tuning across various tasks specified by instructions, the language model demonstrates its proficiency in accurately executing tasks described in previously unseen instructions. However, in black-box distillation, knowledge transfer relies solely on datasets, making the availability of a sufficiently large dataset crucial. Therefore, collaborative efforts in these approaches \cite{wang2022self, peng2023instruction, wu2023lamini, jiang2023lion} involve creating a comprehensive dataset comprising instructions, inputs, and outputs. This dataset enables the student model to acquire extensive knowledge from the teacher model.

Specifically, Wang \etal \cite{wang2022self} propose self-instruction, a semi-automatic process that utilizes indicator signals from the model itself to refine the language model's instructions. The process begins with a constrained seed set of manually crafted tasks, such as the 175 tasks used in our study, to guide the overall generation process. Initially, the prompt model uses this initial set of instructions to generate a broader array of task descriptions. Furthermore, for newly generated sets of instructions, the framework creates input-output instances that can be used for supervised instruction tuning in the future. Finally, various heuristic methods are employed to automatically filter out low-quality or duplicate instructions before incorporating the remaining valid tasks into the task pool. This iterative process can be repeated multiple times until a significant number of tasks are obtained. This method has influenced subsequent research, leading to adjustments in the 13B open-source models like Alpaca \cite{alpaca}, Vicuna \cite{chiang2023vicuna}, and GPT4All \cite{anand2023gpt4all} following this paradigm. Expanding on these ideas, Peng \etal \cite{peng2023instruction} explore the use of GPT-4 to generate instruction-following data for fine-tuning LLMs. They curated a dataset of 52,000 instruction-following examples in both English and Chinese, along with feedback datasets generated by GPT-4. Using these datasets, they fine-tuned two student models, LLaMA-GPT4 and LLaMA-GPT4-CN. Additionally, they developed a feedback model to evaluate the quality of model responses. Wu \etal \cite{wu2023lamini} meticulously compiled a dataset comprising 2.58 million instructions, ensuring coverage of diverse topics. These instructions were used as input to generate responses using GPT-3.5 Turbo. They fine-tuned a range of models under the LaMini-LM, including both encoder-decoder and decoder-only architectures. Evaluation of the LaMini-LM models' performance involved applying automatic metrics across 15 benchmarks, alongside manual assessment. Results illustrate that the proposed LaMini-LM model achieves comparable performance to competitive baselines despite being only one-tenth the size.

However, existing methodologies have predominantly concentrated on one-way knowledge distillation, where student model responses align with those of teacher models to generate instructions without incorporating a "feedback" mechanism. To address this limitation, Jiang \etal \cite{jiang2023lion} introduce an innovative adversarial distillation framework consisting of three stages: imitation, discrimination, and generation. Leveraging the adaptable nature of LLMs, this framework incentivizes teacher models to identify "challenging" instructions and generate new instructions for student models, thereby enhancing the effectiveness of knowledge transfer. This approach achieves open-generation capability comparable to ChatGPT using only 70,000 training samples, surpassing traditional state-of-the-art instruction adjustment models (such as Vicuna-13B) by 55.4\% and 16.7\% on the zero-shot inference BBH and AGIEval tasks, respectively. In efforts to provide task-specific guidance, Chen \etal \cite{chen2023personalized} propose a fine-tuning dataset for code generation instructions and develop a multi-round personalized distillation approach. This approach enables student models to first attempt solving tasks independently, followed by adaptive refinements provided by the teacher to enhance their performance through executive feedback. Unlike traditional knowledge transfer methods where the teacher's prior knowledge is directly imparted to students, personalized refinement offers individualized learning experiences by learning solely from examples of mistakes and iteratively improving their solutions. Meanwhile, UniversalNER \cite{zhou2023universalner} has conducted extensive research on named entity recognition tasks. Unlike the aforementioned methods that aim to increase instruction diversity, UniversalNER focuses on augmenting input diversity to enhance the model's generalization capabilities across various domains.

\begin{table}
\caption{The evaluation results of GPT-2.}
\label{tab0000}
\resizebox{\linewidth}{!}{
\begin{tabular}{@{}c|l|ccccccc|cc@{}}
\toprule
\multirow{2}{*}{Params} & \multicolumn{1}{c|}{\multirow{2}{*}{Method}} & \multicolumn{7}{c|}{Adversarial Robustness(ASR↓)}                                                                                                                                     & \multicolumn{2}{c}{OOD Robustness(F1↑)} \\ \cmidrule(l){3-11} 
                        & \multicolumn{1}{c|}{}                        & \multicolumn{1}{c|}{SST-2} & \multicolumn{1}{c|}{QQP}   & \multicolumn{1}{c|}{MNLI}  & \multicolumn{1}{c|}{QNLI}  & \multicolumn{1}{c|}{RTE}   & \multicolumn{1}{c|}{MNLI-MM} & ANLI  & \multicolumn{1}{c|}{Flipkart} & DDXPlus \\ \midrule
1.5B                    & Teacher                                      & \multicolumn{1}{c|}{62.84} & \multicolumn{1}{c|}{94.87} & \multicolumn{1}{c|}{76.03} & \multicolumn{1}{c|}{75.68} & \multicolumn{1}{c|}{56.79} & \multicolumn{1}{c|}{74.07}   & 89.42 & \multicolumn{1}{c|}{10.72}    & 0       \\ \midrule
\multirow{4}{*}{120M}   & ANLI                                         & \multicolumn{1}{c|}{$\textbf{56.76}^{\ast}$} & \multicolumn{1}{c|}{74.36} & \multicolumn{1}{c|}{73.55} & \multicolumn{1}{c|}{$\textbf{57.43}^{\ast}$} & \multicolumn{1}{c|}{$\textbf{45.68}^{\ast}$} & \multicolumn{1}{c|}{$\textbf{64.81}^{\ast}$}   & 92.75 & \multicolumn{1}{c|}{$\textbf{25.32}^{\ast}$}    & 0       \\
                        & CQA                                          & \multicolumn{1}{c|}{85.81} & \multicolumn{1}{c|}{96.15} & \multicolumn{1}{c|}{95.87} & \multicolumn{1}{c|}{96.62} & \multicolumn{1}{c|}{98.77} & \multicolumn{1}{c|}{93.83}   & 94.17 & \multicolumn{1}{c|}{8.56}     & 0       \\
                        & e-SNLI                                       & \multicolumn{1}{c|}{99.32} & \multicolumn{1}{c|}{$\textbf{66.67}^{\ast}$} & \multicolumn{1}{c|}{$\textbf{65.29}^{\ast}$} & \multicolumn{1}{c|}{60.81} & \multicolumn{1}{c|}{58.02} & \multicolumn{1}{c|}{71.60}   & $\textbf{68.50}^{\ast}$ & \multicolumn{1}{c|}{1.38}     & 0       \\
                        & SVAMP                                        & \multicolumn{1}{c|}{80.41} & \multicolumn{1}{c|}{93.59} & \multicolumn{1}{c|}{85.95} & \multicolumn{1}{c|}{90.54} & \multicolumn{1}{c|}{93.83} & \multicolumn{1}{c|}{82.72}   & 95.42 & \multicolumn{1}{c|}{3.98}     & $\textbf{1.02}^{\ast}$    \\ \midrule
\multirow{4}{*}{340M}   & ANLI                                         & \multicolumn{1}{c|}{58.78} & \multicolumn{1}{c|}{$\textbf{50.00}^{\ast}$} & \multicolumn{1}{c|}{$\textbf{78.51}$} & \multicolumn{1}{c|}{$\textbf{61.49}^{\ast}$} & \multicolumn{1}{c|}{46.91} & \multicolumn{1}{c|}{68.52}   & 87.08 & \multicolumn{1}{c|}{27.39}    & 0       \\
                        & CQA                                          & \multicolumn{1}{c|}{52.70} & \multicolumn{1}{c|}{87.18} & \multicolumn{1}{c|}{88.43} & \multicolumn{1}{c|}{94.59} & \multicolumn{1}{c|}{97.53} & \multicolumn{1}{c|}{91.36}   & 94.08 & \multicolumn{1}{c|}{$\textbf{31.82}^{\ast}$}    & 0       \\
                        & e-SNLI                                       & \multicolumn{1}{c|}{99.32} & \multicolumn{1}{c|}{71.79} & \multicolumn{1}{c|}{80.99} & \multicolumn{1}{c|}{65.54} & \multicolumn{1}{c|}{$\textbf{41.98}^{\ast}$} & \multicolumn{1}{c|}{$\textbf{66.67}^{\ast}$}   & $\textbf{72.83}^{\ast}$ & \multicolumn{1}{c|}{4.61}     & 0       \\
                        & SVAMP                                        & \multicolumn{1}{c|}{$\textbf{50.00}^{\ast}$} & \multicolumn{1}{c|}{69.23} & \multicolumn{1}{c|}{80.17} & \multicolumn{1}{c|}{77.03} & \multicolumn{1}{c|}{64.20} & \multicolumn{1}{c|}{77.78}   & 76.08 & \multicolumn{1}{c|}{25.60}    & 0       \\ \midrule
\multirow{4}{*}{760M}   & ANLI                                         & \multicolumn{1}{c|}{89.19} & \multicolumn{1}{c|}{89.74} & \multicolumn{1}{c|}{88.43} & \multicolumn{1}{c|}{75.68} & \multicolumn{1}{c|}{93.83} & \multicolumn{1}{c|}{86.42}   & \textbf{92.83} & \multicolumn{1}{c|}{6.48}     & 0       \\
                        & CQA                                          & \multicolumn{1}{c|}{$\textbf{54.73}^{\ast}$} & \multicolumn{1}{c|}{$\textbf{83.33}^{\ast}$} & \multicolumn{1}{c|}{90.08} & \multicolumn{1}{c|}{75.00} & \multicolumn{1}{c|}{\textbf{83.95}} & \multicolumn{1}{c|}{94.44}   & 98.83 & \multicolumn{1}{c|}{$\textbf{33.03}^{\ast}$}    & 0       \\
                        & e-SNLI                                       & \multicolumn{1}{c|}{100}   & \multicolumn{1}{c|}{100}   & \multicolumn{1}{c|}{100}   & \multicolumn{1}{c|}{100}   & \multicolumn{1}{c|}{100}   & \multicolumn{1}{c|}{100}     & 100   & \multicolumn{1}{c|}{1.38}     & 0       \\
                        & SVAMP                                        & \multicolumn{1}{c|}{64.19} & \multicolumn{1}{c|}{70.51} & \multicolumn{1}{c|}{\textbf{87.60}} & \multicolumn{1}{c|}{$\textbf{70.27}^{\ast}$} & \multicolumn{1}{c|}{85.19} & \multicolumn{1}{c|}{\textbf{85.19}}   & 96.75 & \multicolumn{1}{c|}{9.84}     & 0       \\ \bottomrule
\end{tabular}
}
\end{table}

\begin{table}
\caption{The evaluation results of OPT.}
\label{tab00000}
\resizebox{\linewidth}{!}{
\begin{tabular}{@{}c|l|ccccccc|cc@{}}
\toprule
\multirow{2}{*}{Params} & \multicolumn{1}{c|}{\multirow{2}{*}{Method}} & \multicolumn{7}{c|}{Adversarial Robustness(ASR↓)}                                                                                                                                     & \multicolumn{2}{c}{OOD Robustness(F1↑)} \\ \cmidrule(l){3-11} 
                        & \multicolumn{1}{c|}{}                        & \multicolumn{1}{c|}{SST-2} & \multicolumn{1}{c|}{QQP}   & \multicolumn{1}{c|}{MNLI}  & \multicolumn{1}{c|}{QNLI}  & \multicolumn{1}{c|}{RTE}   & \multicolumn{1}{c|}{MNLI-MM} & ANLI  & \multicolumn{1}{c|}{Flipkart} & DDXPlus \\ \midrule
13B                     & Teacher                                      & \multicolumn{1}{c|}{68.24} & \multicolumn{1}{c|}{80.77} & \multicolumn{1}{c|}{74.38} & \multicolumn{1}{c|}{50.68} & \multicolumn{1}{c|}{56.79} & \multicolumn{1}{c|}{69.75}   & 72.33 & \multicolumn{1}{c|}{32.04}    & 0       \\ \midrule
\multirow{4}{*}{1.3B}   & ANLI                                         & \multicolumn{1}{c|}{100}   & \multicolumn{1}{c|}{100}   & \multicolumn{1}{c|}{100}   & \multicolumn{1}{c|}{100}   & \multicolumn{1}{c|}{100}   & \multicolumn{1}{c|}{100}     & 100   & \multicolumn{1}{c|}{0}        & 0       \\
                        & CQA                                          & \multicolumn{1}{c|}{50.00} & \multicolumn{1}{c|}{\textbf{85.90}} & \multicolumn{1}{c|}{$\textbf{57.85}^{\ast}$} & \multicolumn{1}{c|}{\textbf{66.89}} & \multicolumn{1}{c|}{$\textbf{43.21}^{\ast}$} & \multicolumn{1}{c|}{$\textbf{64.81}^{\ast}$}   & \textbf{96.00} & \multicolumn{1}{c|}{36.00}    & 0       \\
                        & e-SNLI                                       & \multicolumn{1}{c|}{75.00} & \multicolumn{1}{c|}{93.59} & \multicolumn{1}{c|}{81.82} & \multicolumn{1}{c|}{83.78} & \multicolumn{1}{c|}{67.90} & \multicolumn{1}{c|}{79.63}   & 97.33 & \multicolumn{1}{c|}{2.93}     & 0       \\
                        & SVAMP                                        & \multicolumn{1}{c|}{$\textbf{49.32}^{\ast}$} & \multicolumn{1}{c|}{92.31} & \multicolumn{1}{c|}{78.51} & \multicolumn{1}{c|}{78.38} & \multicolumn{1}{c|}{53.09} & \multicolumn{1}{c|}{75.93}   & 99.33 & \multicolumn{1}{c|}{$\textbf{52.44}^{\ast}$}    & $\textbf{0.85}^{\ast}$    \\ \midrule
\multirow{4}{*}{2.7B}   & ANLI                                         & \multicolumn{1}{c|}{100}   & \multicolumn{1}{c|}{100}   & \multicolumn{1}{c|}{100}   & \multicolumn{1}{c|}{100}   & \multicolumn{1}{c|}{100}   & \multicolumn{1}{c|}{97.53}   & 98.50 & \multicolumn{1}{c|}{0}        & 0       \\
                        & CQA                                          & \multicolumn{1}{c|}{$\textbf{50.68}^{\ast}$} & \multicolumn{1}{c|}{87.18} & \multicolumn{1}{c|}{80.99} & \multicolumn{1}{c|}{75.00} & \multicolumn{1}{c|}{$\textbf{38.27}^{\ast}$} & \multicolumn{1}{c|}{$\textbf{61.11}^{\ast}$}   & 81.92 & \multicolumn{1}{c|}{31.15}    & $\textbf{1.33}^{\ast}$    \\
                        & e-SNLI                                       & \multicolumn{1}{c|}{87.84} & \multicolumn{1}{c|}{79.49} & \multicolumn{1}{c|}{90.91} & \multicolumn{1}{c|}{87.84} & \multicolumn{1}{c|}{86.42} & \multicolumn{1}{c|}{86.42}   & 96.50 & \multicolumn{1}{c|}{3.48}     & 0       \\
                        & SVAMP                                        & \multicolumn{1}{c|}{51.35} & \multicolumn{1}{c|}{$\textbf{65.38}^{\ast}$} & \multicolumn{1}{c|}{$\textbf{68.60}^{\ast}$} & \multicolumn{1}{c|}{\textbf{61.49}} & \multicolumn{1}{c|}{66.67} & \multicolumn{1}{c|}{64.20}   & $\textbf{70.33}^{\ast}$ & \multicolumn{1}{c|}{$\textbf{43.51}^{\ast}$}    & 0.08    \\ \midrule
\multirow{4}{*}{6.7B}   & ANLI                                         & \multicolumn{1}{c|}{100}   & \multicolumn{1}{c|}{100}   & \multicolumn{1}{c|}{99.17} & \multicolumn{1}{c|}{100}   & \multicolumn{1}{c|}{100}   & \multicolumn{1}{c|}{100}     & 99.83 & \multicolumn{1}{c|}{0}        & 0       \\
                        & CQA                                          & \multicolumn{1}{c|}{$\textbf{52.03}^{\ast}$} & \multicolumn{1}{c|}{98.72} & \multicolumn{1}{c|}{85.12} & \multicolumn{1}{c|}{\textbf{83.78}} & \multicolumn{1}{c|}{\textbf{64.20}} & \multicolumn{1}{c|}{93.83}   & 97.08 & \multicolumn{1}{c|}{$\textbf{44.62}^{\ast}$}    & 0       \\
                        & e-SNLI                                       & \multicolumn{1}{c|}{100}   & \multicolumn{1}{c|}{$\textbf{67.95}^{\ast}$} & \multicolumn{1}{c|}{91.74} & \multicolumn{1}{c|}{84.46} & \multicolumn{1}{c|}{70.37} & \multicolumn{1}{c|}{85.80}   & 94.75 & \multicolumn{1}{c|}{0.87}     & 0       \\
                        & SVAMP                                        & \multicolumn{1}{c|}{58.78} & \multicolumn{1}{c|}{84.62} & \multicolumn{1}{c|}{$\textbf{76.03}$} & \multicolumn{1}{c|}{94.59} & \multicolumn{1}{c|}{86.42} & \multicolumn{1}{c|}{\textbf{72.84}}   & \textbf{82.83} & \multicolumn{1}{c|}{31.77}    & $\textbf{0.15}^{\ast}$    \\ \bottomrule
\end{tabular}
}
\end{table}

\subsection{Robustness Evaluation of Black-box KD}

Inspired by the work in \cite{hsieh2023distilling}, we conducted a unified evaluation of the step-by-step distillation algorithm based on CoT from a robustness perspective. Due to the closed-source nature of the PaLM 540B model, we adhered to the experimental setup in \cite{hsieh2023distilling} and used the generated CoT interpretations to fine-tune the student model. The experimental results are presented in Tables \ref{tab0000}-\ref{tab000000}. For GPT-2 models with 120M and 340M parameters, distillation using the interpretations from the ANLI and e-SNLI datasets produced better results. However, as the model size increases, the explanatory power of these two datasets diminishes, and a similar trend is observed in OPT models. For OPT models of various sizes, the explanatory distillation effects generated by ANLI and e-SNLI were suboptimal. This suggests that commonsense data (CQA) and mathematical data (SVAMP) are more conducive to CoT distillation in OPT models. Regardless of whether it is LLaMA or OPT, the distillation of CoT using CQA and SVAMP outperforms the distillation using the other two datasets on Flipkart and DDXPlus. This indicates that distillation of mathematical abilities and commonsense knowledge enhances the model's ability to generalize to out-of-distribution.

\begin{table}
\caption{The evaluation results of LLaMA and LLaMA2.}
\label{tab000000}
\resizebox{\linewidth}{!}{
\begin{tabular}{@{}c|l|ccccccc|cc@{}}
\toprule
                         & \multicolumn{1}{c|}{}                         & \multicolumn{7}{c|}{{\color[HTML]{000000} Adversarial Robustness(ASR↓)}}                                                                                                                                                                                                                                                                        & \multicolumn{2}{c}{{\color[HTML]{000000} OOD Robustness(F1↑)}}                        \\ \cmidrule(l){3-11} 
\multirow{-2}{*}{Params} & \multicolumn{1}{c|}{\multirow{-2}{*}{Method}} & \multicolumn{1}{c|}{{\color[HTML]{000000} SST-2}} & \multicolumn{1}{c|}{{\color[HTML]{000000} QQP}} & \multicolumn{1}{c|}{{\color[HTML]{000000} MNLI}} & \multicolumn{1}{c|}{{\color[HTML]{000000} QNLI}} & \multicolumn{1}{c|}{{\color[HTML]{000000} RTE}} & \multicolumn{1}{c|}{{\color[HTML]{000000} MNLI-MM}} & {\color[HTML]{000000} ANLI} & \multicolumn{1}{c|}{{\color[HTML]{000000} Flipkart}} & {\color[HTML]{000000} DDXPlus} \\ \midrule
13B                      & Teacher                                       & \multicolumn{1}{c|}{50.68}                        & \multicolumn{1}{c|}{69.23}                      & \multicolumn{1}{c|}{62.81}                       & \multicolumn{1}{c|}{58.78}                       & \multicolumn{1}{c|}{46.91}                      & \multicolumn{1}{c|}{77.78}                          & 69.83                       & \multicolumn{1}{c|}{49.16}                           & 5.74                           \\ \midrule
                         & ANLI                                          & \multicolumn{1}{c|}{$\textbf{41.22}^{\ast}$}                        & \multicolumn{1}{c|}{60.26}                      & \multicolumn{1}{c|}{$\textbf{53.72}^{\ast}$}                       & \multicolumn{1}{c|}{$\textbf{47.97}^{\ast}$}                       & \multicolumn{1}{c|}{54.32}                      & \multicolumn{1}{c|}{$\textbf{61.11}^{\ast}$}                          & 71.08                       & \multicolumn{1}{c|}{2.16}                            & 4.80                           \\
                         & CQA                                           & \multicolumn{1}{c|}{46.62}                        & \multicolumn{1}{c|}{$\textbf{46.15}^{\ast}$}                      & \multicolumn{1}{c|}{66.12}                       & \multicolumn{1}{c|}{53.38}                       & \multicolumn{1}{c|}{\textbf{46.91}}                      & \multicolumn{1}{c|}{64.81}                          & 77.42                       & \multicolumn{1}{c|}{\textbf{41.37}}                           & $\textbf{8.33}^{\ast}$                           \\
                         & e-SNLI                                        & \multicolumn{1}{c|}{49.32}                        & \multicolumn{1}{c|}{84.62}                      & \multicolumn{1}{c|}{57.85}                       & \multicolumn{1}{c|}{58.11}                       & \multicolumn{1}{c|}{61.73}                      & \multicolumn{1}{c|}{63.58}                          & \textbf{70.33}                       & \multicolumn{1}{c|}{1.38}                            & 1.46                           \\
\multirow{-4}{*}{LLaMA\_7B}     & SVAMP                                         & \multicolumn{1}{c|}{46.62}                        & \multicolumn{1}{c|}{74.36}                      & \multicolumn{1}{c|}{72.73}                       & \multicolumn{1}{c|}{52.70}                       & \multicolumn{1}{c|}{61.73}                      & \multicolumn{1}{c|}{67.90}                          & 77.75                       & \multicolumn{1}{c|}{32.44}                           & 1.46                           \\ \midrule
13B                      & Teacher                                       & \multicolumn{1}{c|}{53.38}                        & \multicolumn{1}{c|}{50.00}                      & \multicolumn{1}{c|}{71.07}                       & \multicolumn{1}{c|}{45.95}                       & \multicolumn{1}{c|}{50.62}                      & \multicolumn{1}{c|}{62.35}                          & 71.83                       & \multicolumn{1}{c|}{33.76}                           & 18.92                          \\ \midrule
                         & ANLI                                          & \multicolumn{1}{c|}{64.19}                        & \multicolumn{1}{c|}{61.54}                      & \multicolumn{1}{c|}{$\textbf{61.98}^{\ast}$}                       & \multicolumn{1}{c|}{52.70}                       & \multicolumn{1}{c|}{56.79}                      & \multicolumn{1}{c|}{61.11}                          & $\textbf{62.08}^{\ast}$                      & \multicolumn{1}{c|}{23.88}                           & 1.02                           \\
                         & CQA                                           & \multicolumn{1}{c|}{$\textbf{44.59}^{\ast}$}                        & \multicolumn{1}{c|}{48.72}                      & \multicolumn{1}{c|}{66.94}                       & \multicolumn{1}{c|}{50.00}                       & \multicolumn{1}{c|}{43.21}                      & \multicolumn{1}{c|}{66.05}                          & 66.50                       & \multicolumn{1}{c|}{$\textbf{35.10}^{\ast}$}                           & 8.19                           \\
                         & e-SNLI                                        & \multicolumn{1}{c|}{61.49}                        & \multicolumn{1}{c|}{$\textbf{44.87}^{\ast}$}                      & \multicolumn{1}{c|}{74.38}                       & \multicolumn{1}{c|}{$\textbf{40.54}^{\ast}$}                       & \multicolumn{1}{c|}{$\textbf{41.98}^{\ast}$}                      & \multicolumn{1}{c|}{$\textbf{58.64}^{\ast}$}                          & 67.33                       & \multicolumn{1}{c|}{5.48}                            & 4.71                           \\
\multirow{-4}{*}{LLaMA2\_7B}     & SVAMP                                         & \multicolumn{1}{c|}{58.11}                        & \multicolumn{1}{c|}{53.85}                      & \multicolumn{1}{c|}{78.51}                       & \multicolumn{1}{c|}{56.08}                       & \multicolumn{1}{c|}{61.73}                      & \multicolumn{1}{c|}{69.14}                          & 70.42                       & \multicolumn{1}{c|}{27.01}                           & $\textbf{11.67}$                          \\ \bottomrule
\end{tabular}
}
\end{table}

\subsection{Discussion on Black-box KD}

The black-box based KD method is typically used by LLMs to generate explanations or instruction pairs to fine-tune the student model. In this approach, only the teacher model generates data, and only the student model is involved in training, making it memory-efficient. However, most current methods rely on closed-source teacher models, and generating additional data can be costly. Additionally, many methods do not have open-source data generation techniques or involve closed-source generated data, posing challenges for the fair evaluation of these black-box based distillation algorithms.

\subsection{Others}

As large language models have advanced significantly, their inherent limitation lies in their inability to comprehend visual information, as they are primarily designed for processing discrete texts. Consequently, researchers are increasingly exploring ways to transfer the capabilities of language models into multimodal domains, where text and image data are integrated to enable a wider range of tasks \cite{yang2023mm, driess2023palm, gong2023multimodal}. Extracting knowledge from pre-trained multimodal models to enhance the performance and generalization of compact multimodal language models has become a focal point of interest in this field.

\subsubsection{Muiti-Modal Large Language Models}

Knowledge distillation for multimodal large models is still in its nascent stages, focusing primarily on refining instruction-following capabilities. Li \etal \cite{li2023unlock} have pioneered a novel framework featuring two stages for distilling knowledge in multimodal large models. The initial stage involves multimodal pre-training to align multimodal features through a projection layer. The second stage, termed multimodal competitive distillation, establishes a bidirectional feedback loop encompassing: 1) Multimodal instruction adjustment to ensure student responses align with teacher-provided multimodal instructions. 2) Multimodal evaluation to identify challenging multimodal instructions. 3) Multimodal augmentation, where new instructions are generated and combined with original images to create a new multimodal instruction dataset for training student models. Evaluation on datasets like ScienceQA \cite{lu2022learn}, SEED-Bench \cite{li2023seed}, and LLaVA Test Set \cite{liu2024visual} demonstrates that CoMD surpasses existing models in inference tasks and zero-shot settings. Park \etal \cite{park2023localized} developed a localized visual commonsense model by sampling localized commonsense knowledge from LLMs. Users can specify regions as inputs, and a separately trained critic model selects high-quality examples. Empirical results and human evaluations in the zero-shot setting indicate that this distillation method produces a more accurate VL inference model compared to simply passing generated reference expressions to baseline LLMs. Similarly, Hu \etal \cite{hu2023visual} introduced Instruction Tuning for Visual Program Distillation (VPD). VPD leverages LLMs' inference capability by sampling multiple candidate programs, executing and verifying them, and translating correct programs into language descriptions of inference steps for VLM distillation. Extensive experiments have shown that VPD enhances counting, spatial relationship understanding, and combinatorial reasoning abilities in VLMs, achieving state-of-the-art performance in challenging visual tasks such as MMBench \cite{liu2023mmbench}, OK-VQA \cite{marino2019ok}, A-OKVQA \cite{schwenk2022okvqa}, TallyQA \cite{acharya2019tallyqa}, POPE \cite{li2023evaluating}, and Hateful Memes \cite{kiela2020hateful}.

\section{Applications}
\label{sec-app}
In this section, we briefly explore the applications of LLM distillation in various critical domains such as healthcare, education, and law.

\begin{table}
\caption{Applications of LLM distillation.}
\label{tab4}
\resizebox{\linewidth}{!}{
\begin{tabular}{@{}llcccc@{}}
\toprule
Models & Distillation Scenario  & Teacher Model  & Compression Rate  & Evaluation Task & Comparison with Teacher Model

\\ \midrule

HuatuoGPT \cite{zhang2023huatuogpt}    & Healthcare  &  GPT-3.5$_{turbo}$  & /  &  cMedQA2\cite{zhang2018multi}/webMedQA\cite{he2019applying}/Huatuo-26M\cite{li2023huatuo}  &  25.1/18.6 (135\% performance)   \\
Chatdoctor \cite{li2023chatdoctor} & Healthcare  &  GPT-3.5$_{turbo}$  & /  &  HealthCareMagic100k\cite{li2023chatdoctor}  &  84.5/84.1 (100\% performance)   \\
PMC-LLaMA \cite{wu2023pmc} & Healthcare  &  ChatGPT  &  / &  PubMedQA\cite{jin2019pubmedqa}/MedMCQA\cite{pal2022medmcqa}/USMLE\cite{jin2021disease}  &  64.4/55.0 (117\% performance)   \\ \midrule
DARWIN \cite{xie2023darwin} & Education  &  GPT-3$_{175B}$  & 25$\times$  & SciQ\cite{welbl2017crowdsourcing}/FAIR\cite{scheffler2022fair} &  93.6/82.7 (113\% performance)  \\
WizardMath \cite{luo2023wizardmath} & Education  &  ChatGPT  & /  & GSM8k\cite{cobbe2021training}/MATH\cite{hendrycks2021measuring} &  52.2/57.5 (91\% performance)  \\
K2 \cite{deng2023k2} & Education  &  LLaMA$_{7B}$  & 1750$\times$  & GeoBench\cite{deng2023k2} &  34.6/24.6 (141\% performance)  \\ \midrule
LawyerLLaMA \cite{huang2023lawyer} & Law  &  GPT-3.5$_{turbo}$  & /  &  C3\cite{sun2020investigating}/CMNLI\cite{xu2020clue}/SciQ\cite{welbl2017crowdsourcing}/PIQA\cite{bisk2020piqa}  &  / \\
ChatLaw \cite{cui2023chatlaw} & Law  &  Ziya-LLaMA$_{13B}$  & /  & Legal Multiple-choice Questions\cite{cui2023chatlaw} &  / \\ \bottomrule
\end{tabular}
}
\end{table}

\subsection{Healthcare}

Healthcare represents a critical domain deeply intertwined with human well-being. Since the inception of ChatGPT, numerous endeavors have endeavored to harness the prowess of ChatGPT and other LLMs in the realm of medicine. For example, Zhang \etal \cite{zhang2023huatuogpt} introduced HuatuoGPT, a specialized LLM designed for medical consultations. By distilling data from ChatGPT and integrating real-world insights from physicians through supervised fine-tuning, HuatuoGPT incorporates a reward model aimed at synergizing the strengths derived from both datasets. Empirical results demonstrate that HuatuoGPT achieves state-of-the-art performance in medical consultations, outperforming GPT-3.5$_{turbo}$ across various metrics evaluated on GPT-4, including manual assessments and medical benchmark datasets. Li \etal \cite{li2023chatdoctor} highlight the scarcity of LLMs specifically tailored to medical domains. Using LLaMA as a developmental and evaluative platform, they explored two enhancement strategies: model fine-tuning and knowledge integration to augment the efficacy of LLMs as medical chatbots. Fine-tuning the dialogue model on a dataset comprising 100K patient physiological dialogues sourced from online medical consultation platforms, their experiments demonstrate that the Chatdoctor model surpasses ChatGPT in terms of accuracy and F1 score. Furthermore, Wu \etal \cite{wu2023pmc} introduced PMC-LLaMA, which amalgamates 4.8M biomedical academic papers and 30K medical textbooks to infuse data-centric knowledge, coupled with exhaustive fine-tuning tailored to specific domain directives. With a modest parameter count of 13B, PMC-LLaMA demonstrates outstanding performance, surpassing ChatGPT across various public medical question answering benchmarks.

\subsection{Education}

Education represents another critical domain where LLMs show significant promise. Current research demonstrates that LLMs can achieve proficiency comparable to students in standardized exams across various mathematical disciplines such as physics and computer science \cite{achiam2023gpt}. Xie \etal \cite{xie2023darwin} introduced DARWIN, a framework aimed at enhancing natural sciences by accelerating and enriching the automation of discovery processes. This approach incorporates the Scientific Instruction Generation (SIG) model, which integrates structured and unstructured scientific knowledge from public datasets and literature. By eliminating the need for manual extraction or domain-specific knowledge graphs, DARWIN achieves state-of-the-art performance across diverse scientific tasks. Luo \etal \cite{luo2023wizardmath} proposed WizardMath, which utilizes the Reinforcement Learning from Evol-Instruct Feedback (RLEIF) technique to enhance the mathematical reasoning capabilities of LLaMA-2 \cite{touvron2023llama}. This method employs math-specific Evol-Instruct to generate diverse mathematical instruction data, subsequently training the Instruction Reward Model (IRM) and the Process Supervised Reward Model (PRM) \cite{yuan2023scaling}. The IRM evaluates the quality of evolutionary instructions, while the PRM receives feedback at each step of the solution process. Through extensive experimentation on two mathematical reasoning benchmarks, GSM8k \cite{cobbe2021training} and MATH \cite{hendrycks2021measuring}, WizardMath significantly outperforms other open-source LLMs. Furthermore, Deng \etal \cite{deng2023k2} introduced K2, a LLM tailored for geoscience, and established the GeoBench, the first geoscience benchmark, to evaluate LLMs within this domain.

\subsection{Law}

Law, a domain rich in professional expertise, has recently adopted LLMs to address various legal tasks, such as legal document analysis \cite{blair2023can} and legal document generation \cite{choi2021chatgpt}. Huang \etal \cite{huang2023lawyer} integrated legal expertise into the continuous training phase of LLaMA by employing carefully designed supervised fine-tuning tasks. These tasks aimed to impart professional skills to the model while mitigating the issue of model-generated illusions. To enhance training, they introduced a retrieval module that extracts relevant legal articles before the model generates responses. Similarly, Cui \etal \cite{cui2023chatlaw} integrated legal-specific data into LLaMA, resulting in the creation of ChatLaw. Concerned with the accuracy of reference retrieval from legal datasets, they developed a hybrid approach combining vector database retrieval and keyword-based retrieval. This approach addresses hallucination concerns and improves accuracy by implementing a self-attention mechanism. This mechanism enhances the ability of large models to correct errors within reference data, thereby improving coherence and augmenting problem-solving proficiency in legal contexts.

\section{Challenges and Future Directions}
\label{4}

\subsection{Unified Evaluation Benchmark}

The existing benchmark for evaluating knowledge distillation primarily falls into four categories: 1) General Language Understanding Evaluation (GLUE) Benchmark \cite{wang2018glue}: This benchmark consists of nine sentence-level classification tasks, including language acceptability \cite{warstadt2019neural}, sentiment analysis \cite{socher2013recursive}, text similarity \cite{cer2017semeval}, entailment detection \cite{dolan2005automatically}, and natural language inference \cite{rajpurkar2016squad}. It is commonly utilized to assess distillation methods employing BERT as the teacher model. 2) Multimodal Multitask Learning Understanding (MMLU) Benchmark \cite{hendrycks2020measuring}: This benchmark serves as a universal evaluation tool for assessing the multitasking knowledge comprehension abilities of LLMs. It covers various domains such as mathematics, computer science, humanities, and social sciences, featuring tasks of varying difficulty levels from basic to advanced. 3) BIG Bench \cite{srivastava2023beyond}: A collaborative effort to create a comprehensive evaluation benchmark that explores the capabilities of existing LLMs across a diverse range of tasks. It includes 204 tasks spanning linguistics, child development, mathematics, common sense reasoning, biology, physics, social prejudice, software development, and more. 4) Human-Evaluated Language Models (HELM) Benchmark \cite{liang2022holistic}: This is a holistic evaluation benchmark comprising 16 core scenarios and 7 indicator categories. It integrates various previously proposed evaluation benchmarks to provide a holistic assessment of LLM performance. These benchmarks collectively cover a wide array of mainstream LLM evaluation tasks. Additionally, there are specialized evaluation benchmarks tailored to specific tasks, such as TyDiQA \cite{clark2020tydi} for evaluating multilingual knowledge utilization and MGSM \cite{shi2022language} for assessing multilingual mathematical reasoning. As large models continue to evolve, evaluation criteria are continually updated, and developing a unified evaluation standard for knowledge distillation remains a promising avenue of research.

\subsection{Advanced Algorithms}

Current methodologies primarily aim to equip student models with specific capabilities. For example, symbolic knowledge distillation \cite{west2022symbolic} leverages LLMs to gather and filter data, extracting high-quality commonsense maps for training commonsense models. Similarly, DISCO \cite{chen2022disco} employs LLMs to acquire counterfactual data, which is then filtered using a large teacher Natural Language Inference model to improve students' proficiency in natural language reasoning tasks. As open-source LLMs continue to evolve, exploring white-box distillation algorithms for LLMs could prove to be an effective approach for integrating multiple capabilities. Furthermore, the current development pace of MLLMs distillation lags behind that of LLMs. Thus, investigating more advanced MLLMs distillation algorithms could facilitate the integration of multiple modalities more effectively.

\subsection{Interpretability}

Stanton \etal \cite{stanton2021does} explore the interpretability of knowledge distillation and introduce the concept of matching degree to enhance its reliability. Their study reveals several significant insights: 1) The relationship between student models' generalization performance and matching degree is not uniformly consistent. Excluding self-distillation, models with the best generalization performance do not always exhibit the highest fidelity. 2) There is a notable correlation between student models' fidelity and the calibration of the distillation process. Although the most faithful student model may not always achieve the highest accuracy, it consistently shows superior calibration. 3) Optimization during the knowledge distillation process is challenging, resulting in lower fidelity. Similarly, in the era of large language models, knowledge distillation faces comparable difficulties. For example, current methods struggle to elucidate how CoT-distillation imparts CoT capability to student language models or to determine the required amount of data for fine-tuning instructions. Therefore, integrating interpretability into the process is crucial for advancing LLM knowledge distillation. This integration not only aids in evaluating model distillation but also enhances the reliability and predictability of models in production

\section{Conclusion}
\label{5}

In this survey, we systematically investigate the knowledge distillation algorithms from three perspectives: methods, evaluation, and application. Compared to smaller models, distillation in larger models faces more challenges. Despite considerable efforts by existing algorithms to tackle these challenges, many still rely on frameworks initially tailored for compressing smaller models, while the challenge of compressing large models still exists. In the future, while ensuring the universality and generalization of LLMs, it becomes imperative to delve deeper into developing more efficient and effective compression algorithms. This survey aims to furnish valuable references, shed light on the current landscape, and advocate for ongoing exploration of this pivotal theme to enable the effective design, learning, and application of various distillation objectives within the teacher-student framework.

\bibliographystyle{ACM-Reference-Format}
\bibliography{sample-base}

\end{document}